\documentclass[10pt,twocolumn,letterpaper]{article}

\usepackage{cvpr}              % camera-ready
\newcommand{\OURS}{\textsc{MM-UAVBench}\xspace}
\newcommand{\worldwideweb}{\raisebox{-1.5pt}{\includegraphics[height=1.05em]{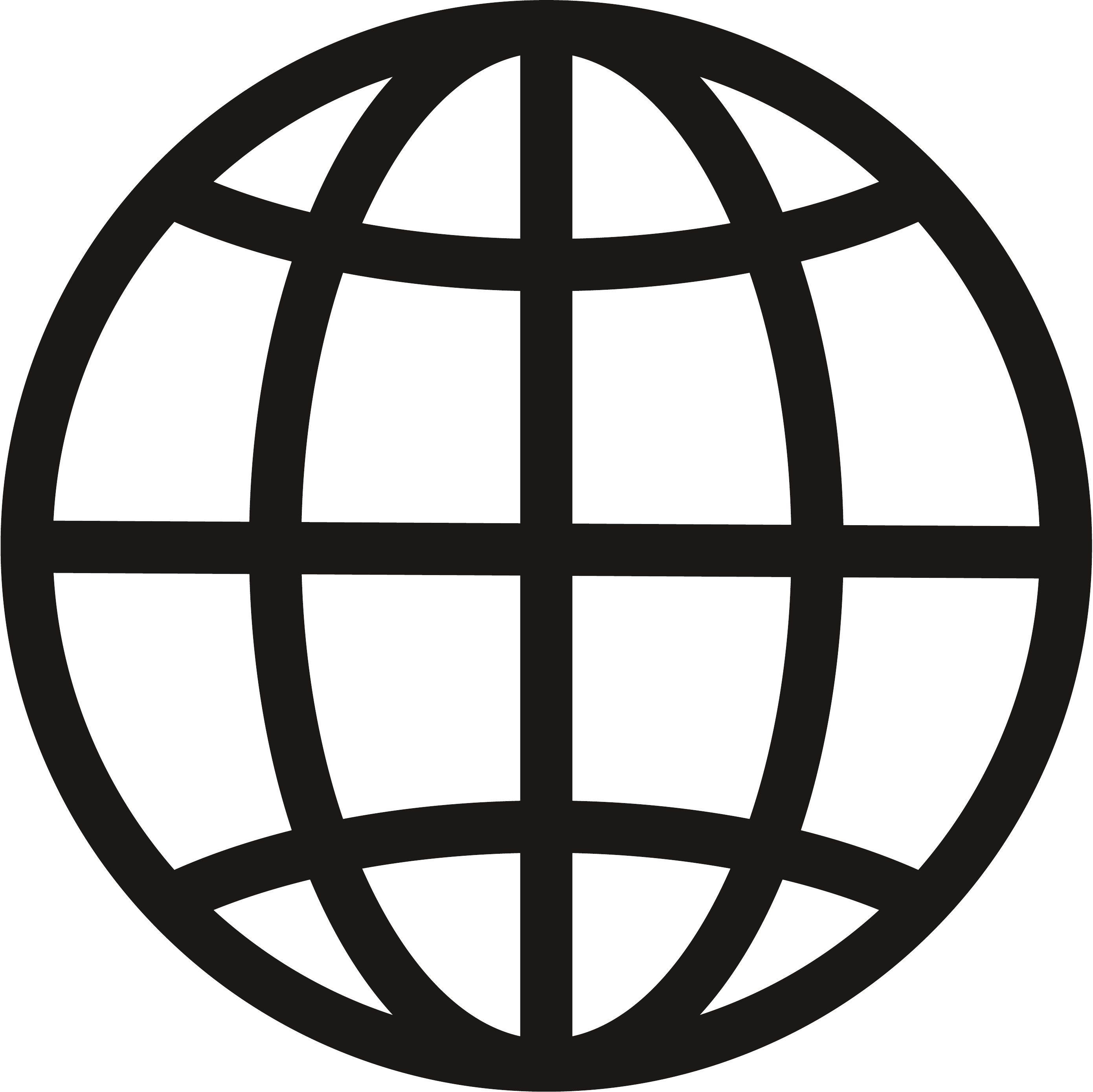}}\xspace}
\newcommand{\github}{\raisebox{-1.5pt}{\includegraphics[height=1.05em]{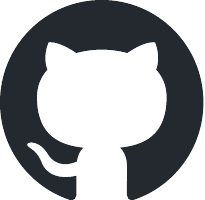}}\xspace}
\newcommand{\huggingface}{\raisebox{-1.5pt}{\includegraphics[height=1.05em]{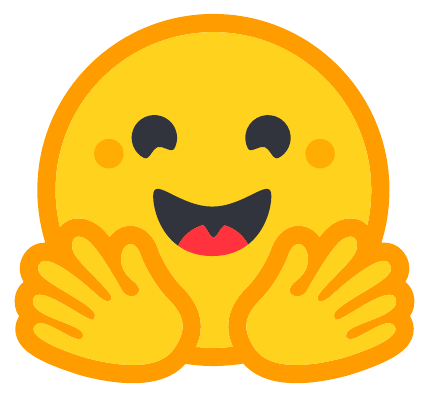}}\xspace}

\usepackage[table]{xcolor}
\usepackage{makecell}
\usepackage{adjustbox}
\usepackage{booktabs}
\usepackage{array}
\usepackage{siunitx}
\usepackage{fp}
\usepackage{microtype}
\usepackage{xcolor} 
\usepackage{multirow}
\usepackage{tcolorbox}
\usepackage{pifont}

\definecolor{indianred}{rgb}{0.8, 0.36, 0.36}
\definecolor{bleudefrance}{rgb}{0.19, 0.55, 0.91}
\definecolor{forestgreen}{rgb}{0.0, 0.5, 0.0}
\definecolor{ashgrey}{rgb}{0.7, 0.75, 0.71}
\definecolor{darkorange}{rgb}{1.0, 0.55, 0.0}
\definecolor{darkorchid}{rgb}{0.6, 0.2, 0.8}

\usepackage{fontawesome5} 
\newcommand{\icoyes}{\textcolor{forestgreen}{\faCheckCircle}\xspace}
\newcommand{\icono}{\textcolor{ashgrey}{\faTimesCircle}\xspace}
\newcommand{\icohalf}{\textcolor{darkorange}{\ding{51}\kern-0.65em\ding{55}}}

\newcommand{\iconurban}{\textcolor{teal}{\faBuilding}\xspace}

\newcommand{\iconnature}{\textcolor{olive}{\faTree}\xspace}

\newcommand{\iconwild}{\textcolor{brown}{\faPaw}\xspace}

\newcommand{\icondisaster}{\textcolor{red!70!black}{\faExclamationTriangle}\xspace}

\newcommand{\iconagri}{\textcolor{green!60!black}{\faLeaf}\xspace}

\usepackage[dvipsnames]{xcolor}  % 允许使用 BrickRed, ForestGreen 等颜色

\newcommand{\poscolor}[1]{\textcolor{ForestGreen}{#1}}
\newcommand{\negcolor}[1]{\textcolor{BrickRed}{#1}}

\newlength{\NumW}
\newcolumntype{N}{>{\centering\arraybackslash}m{\NumW}} % 数字列：定宽+居中+垂直居中
\newcolumntype{L}[1]{>{\centering\arraybackslash}m{#1}} % 通用定宽列：居中+垂直居中

\newlength{\HdrH}
\newlength{\HdrW}

\newcommand{\pcnum}[1]{%
  \FPeval\pcres{round(#1*100,2)}%
  \num{\pcres}% 注意：\num 的格式在表内部单独 \sisetup
}

\newcommand{\narrowbox}[1]{\scalebox{0.6}[0.7]{#1}}

\newcommand{\digitsize}{\fontsize{3}{6}\selectfont}
\newcommand{\CellNum}[1]{\narrowbox{{\digitsize #1}}}

\newcommand{\na}{\textcolor{gray}{--}}
\newcommand{\cellcolorapply}[1]{%
  \begingroup
  \edef\val{#1}%
  \def\NA{\na}%
  \ifx\val\NA
    \na
  \else
    \ifdim\val pt<0.05pt  \cellcolor{blue!85!black}{\textcolor{white}{\CellNum{\pcnum{\val}}}}%
    \else\ifdim\val pt<0.10pt \cellcolor{blue!75}{\textcolor{white}{\CellNum{\pcnum{\val}}}}%
    \else\ifdim\val pt<0.15pt \cellcolor{cyan!80!blue}{\textcolor{black}{\CellNum{\pcnum{\val}}}}%
    \else\ifdim\val pt<0.20pt \cellcolor{cyan!65!white}{\textcolor{black}{\CellNum{\pcnum{\val}}}}%
    \else\ifdim\val pt<0.25pt \cellcolor{cyan!50!white}{\textcolor{black}{\CellNum{\pcnum{\val}}}}%
    \else\ifdim\val pt<0.30pt \cellcolor{cyan!40!yellow!20!white}{\textcolor{black}{\CellNum{\pcnum{\val}}}}%
    \else\ifdim\val pt<0.35pt \cellcolor{yellow!30!cyan!30!white}{\textcolor{black}{\CellNum{\pcnum{\val}}}}%
    \else\ifdim\val pt<0.40pt \cellcolor{yellow!45!white}{\textcolor{black}{\CellNum{\pcnum{\val}}}}%
    \else\ifdim\val pt<0.45pt \cellcolor{yellow!60!white}{\textcolor{black}{\CellNum{\pcnum{\val}}}}%
    \else\ifdim\val pt<0.50pt \cellcolor{yellow}{\textcolor{black}{\CellNum{\pcnum{\val}}}}%
    \else\ifdim\val pt<0.55pt \cellcolor{red!10!yellow}{\textcolor{black}{\CellNum{\pcnum{\val}}}}%
    \else\ifdim\val pt<0.60pt \cellcolor{red!18!yellow}{\textcolor{black}{\CellNum{\pcnum{\val}}}}%
    \else\ifdim\val pt<0.65pt \cellcolor{red!26!yellow}{\textcolor{black}{\CellNum{\pcnum{\val}}}}%
    \else\ifdim\val pt<0.70pt \cellcolor{red!35!yellow}{\textcolor{black}{\CellNum{\pcnum{\val}}}}%
    \else\ifdim\val pt<0.75pt \cellcolor{red!45!yellow}{\textcolor{black}{\CellNum{\pcnum{\val}}}}%
    \else\ifdim\val pt<0.80pt \cellcolor{red!58!yellow}{\textcolor{black}{\CellNum{\pcnum{\val}}}}%
    \else\ifdim\val pt<0.85pt \cellcolor{red!70!yellow}{\textcolor{black}{\CellNum{\pcnum{\val}}}}%
    \else\ifdim\val pt<0.90pt \cellcolor{red!80!yellow}{\textcolor{black}{\CellNum{\pcnum{\val}}}}%
    \else\ifdim\val pt<0.95pt \cellcolor{red!88!yellow}{\textcolor{black}{\CellNum{\pcnum{\val}}}}%
    \else                  \cellcolor{red!95!yellow}{\textcolor{black}{\CellNum{\pcnum{\val}}}}%
    \fi\fi\fi\fi\fi\fi\fi\fi\fi\fi
    \fi\fi\fi\fi\fi\fi\fi\fi\fi
  \fi
  \endgroup
}

\definecolor{cvprblue}{rgb}{0.21,0.49,0.74}
\usepackage[pagebackref,breaklinks,colorlinks,allcolors=cvprblue]{hyperref}

\def\confName{CVPR}
\def\confYear{2026}

\title{\OURS: How Well Do Multimodal Large Language Models\\ See, Think, and Plan in Low-Altitude UAV Scenarios?}
\author{
Shiqi Dai\textsuperscript{1}\thanks{Equal contribution.} \quad
Zizhi Ma\textsuperscript{2}\footnotemark[1] \quad
Zhicong Luo\textsuperscript{3} \quad
Xuesong Yang\textsuperscript{4} \quad
Yibin Huang\textsuperscript{5} \quad
Wanyue Zhang\textsuperscript{4} \quad \\
Chi Chen\textsuperscript{1}\thanks{Corresponding author.}  \quad
Zonghao Guo\textsuperscript{1} \quad
Wang Xu\textsuperscript{1} \quad
Yufei Sun\textsuperscript{2} \quad
Maosong Sun\textsuperscript{1}\footnotemark[2] \\[4pt]
\textsuperscript{1}Tsinghua University \quad
\textsuperscript{2}Nankai University \quad 
\textsuperscript{3}Northwest Polytechnical University \quad \\
\textsuperscript{4}Chinese Academy of Sciences \quad 
\textsuperscript{5}Harbin Institute of Technology \\[2pt]
{\tt\small \{daisq99@gmail.com, chenchithu@gmail.com\}} \\
{\worldwideweb \href{https://mm-uavbench.github.io/}{{\text{Project Page}}}} \quad \quad {\github \href{https://github.com/MM-UAVBench/MM-UAVBench}{{\text{Evaluation Code}}}}
\quad \quad
{\huggingface \href{https://huggingface.co/datasets/daisq/MM-UAVBench}{{\text{\OURS}}}}
}

\begin{document}
\maketitle

% ---- Teaser figure (封面大图) ----
% 单栏版本（更稳）：figure；想跨两栏用 figure* 并把 width=\textwidth
\begin{figure*}[t]
  \centering
  \includegraphics[width=\textwidth]{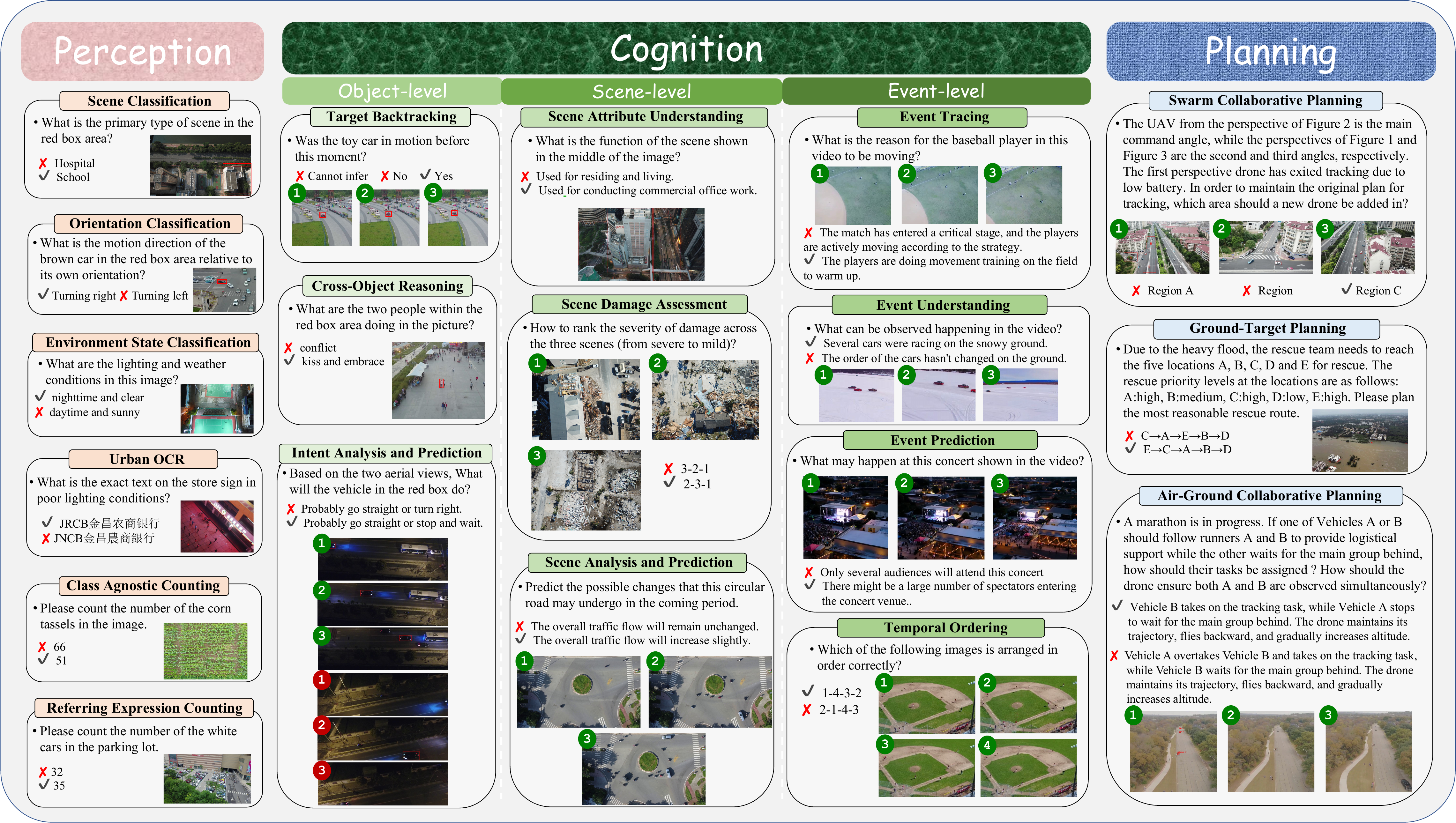} 
  \caption{Overview of \OURS. \OURS consists of 19 tasks covering three core capability dimensions: Perception, Cognition, and Planning. Perception tasks assess basic visual understanding such as classification, OCR, and counting. Cognition tasks span three hierarchical levels—object-level, scene-level, and event-level—evaluating the model’s ability to infer intentions, reason across objects, analyze scenes, understand events, and predict outcomes. Planning tasks assess UAV-specific decision making, including planning for single or multi-UAV systems, directing ground-target actions from an aerial perspective, and coordinating cooperative actions between aerial agents and ground participants. All examples shown are real UAV imagery, illustrating the diverse challenges present in low-altitude scenarios.}
  \label{fig:teaser}
\end{figure*}

% ===== Your sections =====
\begin{abstract}
While Multimodal Large Language Models (MLLMs) have exhibited remarkable general intelligence across diverse domains, their potential in low-altitude applications dominated by Unmanned Aerial Vehicles (UAVs) remains largely underexplored. 
Existing MLLM benchmarks rarely cover the unique challenges of low-altitude scenarios, while UAV-related evaluations mainly focus on specific tasks such as localization or navigation, without a unified evaluation of MLLMs’ general intelligence.
To bridge this gap, we present \OURS, a comprehensive benchmark that systematically evaluates MLLMs across three core capability dimensions—perception, cognition, and planning—in low-altitude UAV scenarios. \OURS comprises 19 sub-tasks with over 5.7K manually annotated questions, all derived from real-world UAV data collected from public datasets.
Extensive experiments on 16 open-source and proprietary MLLMs reveal that current models struggle to adapt to the complex visual and cognitive demands of low-altitude scenarios. Our analyses further uncover critical bottlenecks such as spatial bias and multi-view understanding that hinder the effective deployment of MLLMs in UAV scenarios. We hope \OURS will foster future research on robust and reliable MLLMs for real-world UAV intelligence.

% While numerous benchmarks have demonstrated the general intelligence of Multimodal Large Language Models (MLLMs) across diverse domains, their potential in low-altitude applications dominated by Unmanned Aerial Vehicles (UAVs) remains largely underexplored. 
% Generic MLLM benchmarks are not designed to handle the complexities of low-altitude UAV scenarios, such as environmental monitoring, joint reasoning with ground views, and collaborative tracking with other UAVs. 
% Existing UAV-oriented benchmarks mainly focus on specific low-altitude applications—such as localization, navigation, and anomaly detection—rather than providing a systematic evaluation of MLLMs’ perception, cognition, and planning abilities.
% To bridge this gap, we present UAVBench, a comprehensive MLLM benchmark tailored for low-altitude UAV scenarios. It comprises 19 sub-tasks with over 5.6K multiple-choice questions, all manually annotated and derived from real-world UAV data collected in public datasets. 
% We conduct extensive experiments on UAVBench using 17 open-source and proprietary MLLMs, and further analyze their performance with respect to object scale sensitivity, spatial perception bias, multi-view understanding, and egocentric planning. The results on UAVBench reveal that current MLLMs struggle to adapt to low-altitude tasks, highlighting the necessity of UAV-tailored MLLM designs for effective real-world deployment.

\end{abstract}
\section{Introduction}

With the rapid progress of Multimodal Large Language Models (MLLMs), their capabilities have become increasingly comprehensive~\cite{bai2025qwen25vl,zhu2025internvl3,yao2025minicpmv}. Such integrated intelligence is especially appealing for Unmanned Aerial Vehicles (UAVs), which are evolving from passive sensing platforms into autonomous edge agents in complex low-altitude environments. Integrating MLLMs into UAVs can elevate their intelligence from basic perception-such as object detection~\cite{sun2025refdrone} and tracking~\cite{zhu2021detection}—to cognitive reasoning and task planning, marking a key step toward autonomous aerial intelligence in real-world missions.

Despite this potential, the fundamental abilities of MLLMs to operate or assist in complex low-altitude environments remain largely unevaluated. Most existing benchmarks for evaluating MLLMs focus on general image or video understanding in everyday scenes~\cite{fu2023mme,fu2024video_mme,li2024seedbench,wang2025hrbench}, emphasizing static perception from ground-level or object-centric views. Even when low-altitude imagery occasionally appears in these datasets~\cite{wang2025hrbench}, it is not treated as a distinct evaluation domain.
Meanwhile, several remote sensing benchmarks~\cite{li2024vrsbench,danish2025geobenchvlm,wang2025xlrsbench} assess MLLMs from satellite or aerial perspectives, but they mainly involve high-altitude top-down views with stable geometry and coarse spatial resolution. Consequently, none of these datasets capture the dynamic, near-ground, and multi-agent characteristics inherent to low-altitude UAV scenarios.

Recently, several studies have begun to investigate the applicability of MLLMs in UAV-related scenarios~\cite{du2018unmanned,sun2025refdrone,mo2025a2seek,li2021uav,bozcan2020air,zhao2025urbanvideobench}. However, most of these efforts primarily focus on traditional perception tasks such as object detection~\cite{du2018unmanned,zhu2018vision,bozcan2020air}, referring grounding~\cite{sun2025refdrone}, counting~\cite{hsieh2017drone, wen2021detection} and target tracking~\cite{naik2024bucktales}. Another line of evaluation focuses on navigation and control-oriented tasks, including trajectory following and path planning for UAVs~\cite{zhao2025urbanvideobench,wang2024towards,xiao2025uav}, aiming to assess models’ low-level motion understanding or decision execution. Although these benchmarks contribute to evaluating UAV perception and navigation, they remain task-specific and lack a comprehensive assessment of MLLMs’ higher-level abilities in realistic low-altitude environments.

In realistic UAV operations, intelligence involves more than recognizing objects or following trajectories—it demands understanding what is happening within a scene, how multiple entities (including UAVs and ground targets) interact, and what strategic decisions should follow~\cite{tian2025uavsmeetllms}. \textbf{A systematic evaluation benchmark is needed to measure how well MLLMs see, think, and plan in complex real-world UAV scenarios}.

In this work, we introduce \textbf{\OURS}, a comprehensive benchmark designed to evaluate the perception, cognition, and planning abilities of MLLMs in low-altitude UAV scenarios. \OURS provides a unified evaluation paradigm that reflects the hierarchical intelligence required for real-world aerial missions. It features three main characteristics:

\begin{itemize}
    \item \textbf{Comprehensive Task Design.} \OURS includes 19 tasks across three key capability dimensions and incorporates UAV-specific considerations, including multi-level cognition (object, scene, and event) and planning that involves both aerial and ground agents, resulting in a comprehensive task design.
    \item \textbf{Diverse Real-World Scenarios.} Unlike previous benchmarks that focus on limited scenes or rely on simulated environments, \OURS is constructed from real UAV imagery collected across a wide range of scenarios, including but not limited to urban areas, agricultural fields, wildlife habitats, and emergency or disaster zones, enabling robust and generalizable evaluation.
    \item \textbf{High-quality Human Annotations.} All tasks are manually annotated to ensure both labeling quality and appropriate task difficulty. In addition, we provide multiple forms of detailed auxiliary annotations, such as bounding boxes for key entities, to support in-depth capability analysis.
\end{itemize}

\begin{table*}[t]
  \centering
  \caption{Overview of \OURS and comparison with representative existing benchmarks. \icoyes, \icono, and \icohalf respectively denote datasets constructed from real imagery, purely simulation, or partially real data that include simulated components. The ``Anno.'' column specifies the annotation method of each benchmark,
  including \textit{Human} (purely human-labeled), \textit{Auto} (fully automatic generation), and \textit{Semi-Auto} (generated labels with
  human refinement). Scenario icons: \iconurban~Urban, \iconnature~Natural scenes, 
  \iconwild~Wildlife, 
  \icondisaster~Disaster/Emergency, \iconagri~Agriculture.}
  \small
  \setlength{\tabcolsep}{6pt}
  \renewcommand{\arraystretch}{1.15}
  \begin{adjustbox}{max width=\textwidth}
  \begin{tabular}{@{}l c c c c c c c@{}}
    \toprule
    \textbf{Benchmark} &
    \textbf{Capability Types} &
    \textbf{\#Tasks} &
    \textbf{Scenarios} &
    \textbf{Real Imagery} &
    \textbf{Anno.} &
    \textbf{\#Source} &
    \textbf{\#Test Instances} \\
    \midrule
    \rowcolor{gray!12}
    \multicolumn{7}{l}{\textbf{Remote Sensing Perspective}} \\

    VRSBench~\cite{li2024vrsbench} &
    Perception / Cognition &
    31 &
    -- &
    \icoyes &
    Semi-Auto &
    29.6K images &
    205.3K \\

    XLRS-Bench~\cite{wang2025xlrsbench} &
    Perception / Cognition &
    16 &
    -- &
    \icoyes &
    Semi-Auto &
    1.4K images &
    45.9K \\

    \midrule
    \rowcolor{gray!12}
    \multicolumn{7}{l}{\textbf{UAV Perspective}} \\

    UAVDT~\cite{du2018unmanned} &
    Object Detection and Tracking &
    3 &
    \iconurban &
    \icoyes &
    Human &
    80K images &
    841.5K \\
    
    RefDrone~\cite{sun2025refdrone} &
    Visual Grounding &
    1 &
    \iconurban &
    \icoyes &
    Semi-Auto &
    8.5K images &
    63.6K \\

    UAV-Human~\cite{li2021uav} &
    Human Behavior Understanding &
    4 &
    \iconurban &
    \icoyes &
    Human &
    67.4K videos &
    86.0K \\

    UAV-ON ~\cite{xiao2025uav} &
    Visual-Language Navigation &
    1 &
    \iconurban \iconnature &
    \icono &
    Auto &
    1.2K targets &
    1.2K \\

    OpenUAV ~\cite{wang2024towards} &
    Visual-Language Navigation &
    1 &
    \iconurban \iconnature &
    \icono &
    Auto &
    12.1K trajectories &
    12.1K \\

    MME-RealWorld-MO~\cite{liu2025surveillancevqa589k} &
    High-resolution Understanding &
    6 &
    \iconurban &
    \icoyes &
    Human &
    1.6K images &
    2.2K \\

    SkyAgent-Eval~\cite{yao2024aeroverse} &
    Embodied Capability &
    5 &
    \iconurban &
    \icono &
    Human &
    67.4K videos &
    86.0K \\

    UrbanVideo-Bench~\cite{zhao2025urbanvideobench} &
    Embodied Capability &
    16 &
    \iconurban &
    \icohalf &
    Semi-Auto &
    1.5K videos &
    5.2K \\

    \rowcolor{green!8}
    \textbf{\OURS (Ours)} &
    \textbf{Comprehensive Per. / Cog. / Plan.} &
    \textbf{19} &
    \textbf{\iconurban \iconnature \iconwild \icondisaster \iconagri} &
    \textbf{\icoyes} &
    \textbf{Human} &
    \textbf{1.5K videos + 2.8K images} &
    \textbf{5.7K} \\
    \bottomrule
  \end{tabular}
  \end{adjustbox}
  \label{tab:benchmark_comparison}
\end{table*}

To construct this benchmark, we collect real-world UAV videos and images from diverse data sources, encompassing 1549 video clips and 2873 images with an average resolution of $1622\times1033$. Using these data, we manually annotate 16 tasks, while the remaining 3 tasks are generated through rule-based transformation of manually annotated labels, resulting in 5702 multiple-choice QA annotations in total.
We evaluate a broad set of MLLMs on \OURS and find that their perception capabilities in UAV scenarios remain limited, with even more pronounced deficiencies in cognition and planning tasks. Further analyses on object-scale sensitivity, spatial perception bias, multi-view understanding, and egocentric planning indicate that current MLLMs struggle to adapt to low-altitude UAV challenges, underscoring the need for UAV-tailored model designs for practical deployment.
Our main contributions are summarized as follows:
\begin{itemize}
    \item We present \OURS, a new and comprehensive benchmark for evaluating the perception, cognition, and planning capabilities of MLLMs across 19 tasks in low-altitude UAV scenarios.
    \item We construct \OURS from real-world UAV datasets with both manually annotated and rule-converted tasks, resulting in 5702 high-quality annotations that provide strong data authenticity and well-controlled task difficulty.
    \item We benchmark a series of MLLMs on \OURS and provide detailed analyses that expose critical limitations, highlighting the need for UAV-oriented MLLM designs for real-world deployment.
\end{itemize}

\section{Related Work}

% 逻辑顺序：1. MLLMs发展很好（引用典型MLLM）2. MLLM也应用到空中领域，主要是在遥感上（遥感相关MLLM）3. MLLM在低空也在特定任务上有模型（低空MLLM）4. 目前没有通用的低空MLLM以及评测。低空加一些VLN的benchmark geobench vlm xlrsbench urbanvideobench 

\subsection{General MLLM Benchmark}

A wide range of benchmarks have been developed to evaluate the visual and reasoning abilities of MLLMs. General-purpose benchmarks such as MME~\cite{fu2023mme}, SEED-Bench~\cite{li2024seedbench}, and VideoMME~\cite{fu2024video_mme} offer broad assessments of object recognition, commonsense reasoning, and video understanding across everyday scenes. Recently, several works have also explored comprehensive MLLM evaluation in remote sensing, including VRSBench~\cite{li2024vrsbench} and XLRS-Bench~\cite{wang2025xlrsbench}, but these operate primarily on high-altitude, top-down imagery with stable viewpoints.
Despite their breadth, they do not address the distinct characteristics of low-altitude UAV scenarios, such as dynamic viewpoints, large scale variation, multi-entity interactions, and action-oriented decision making, and thus provide limited insight into the operational intelligence required for UAV missions.

\subsection{Evaluation in Low-Altitude UAV Scenarios}

Existing benchmarks for UAV scenarios mainly cover narrow and task-specific capabilities. Many perception-oriented datasets such as UAVDT~\cite{du2018unmanned} and RefDrone~\cite{sun2025refdrone} focus on detection, tracking, or grounding in limited urban scenes. A second line of work, such as UAV-ON~\cite{xiao2025uav} and OpenUAV~\cite{wang2024towards}, evaluates visual-language navigation in simulator environments. Although these benchmarks introduce decision-oriented tasks, they rely heavily on synthetic scenes and address only ego-centric navigation. More recent embodied UAV evaluations, such as SkyAgent-Eval~\cite{yao2024aeroverse} and UrbanVideo-Bench~\cite{zhao2025urbanvideobench}, examine how MLLMs can assist UAVs in scene perception and flight planning. However, they still lack comprehensive assessments of UAV-perspective scene and event understanding, and their heavy reliance on simulator-generated data introduces potential sim-to-real gaps.
Overall, existing UAV benchmarks offer limited capability coverage and restricted scenario diversity, providing only a partial view of the intelligence required for low-altitude UAV operations. In contrast, \OURS jointly evaluates perception, multi-level cognition, and multi-agent planning across diverse real-world UAV scenarios. A detailed comparison is provided in Table~\ref{tab:benchmark_comparison}.

\section{\OURS}
\begin{figure}[t]
    \centering
    \includegraphics[width=\linewidth]{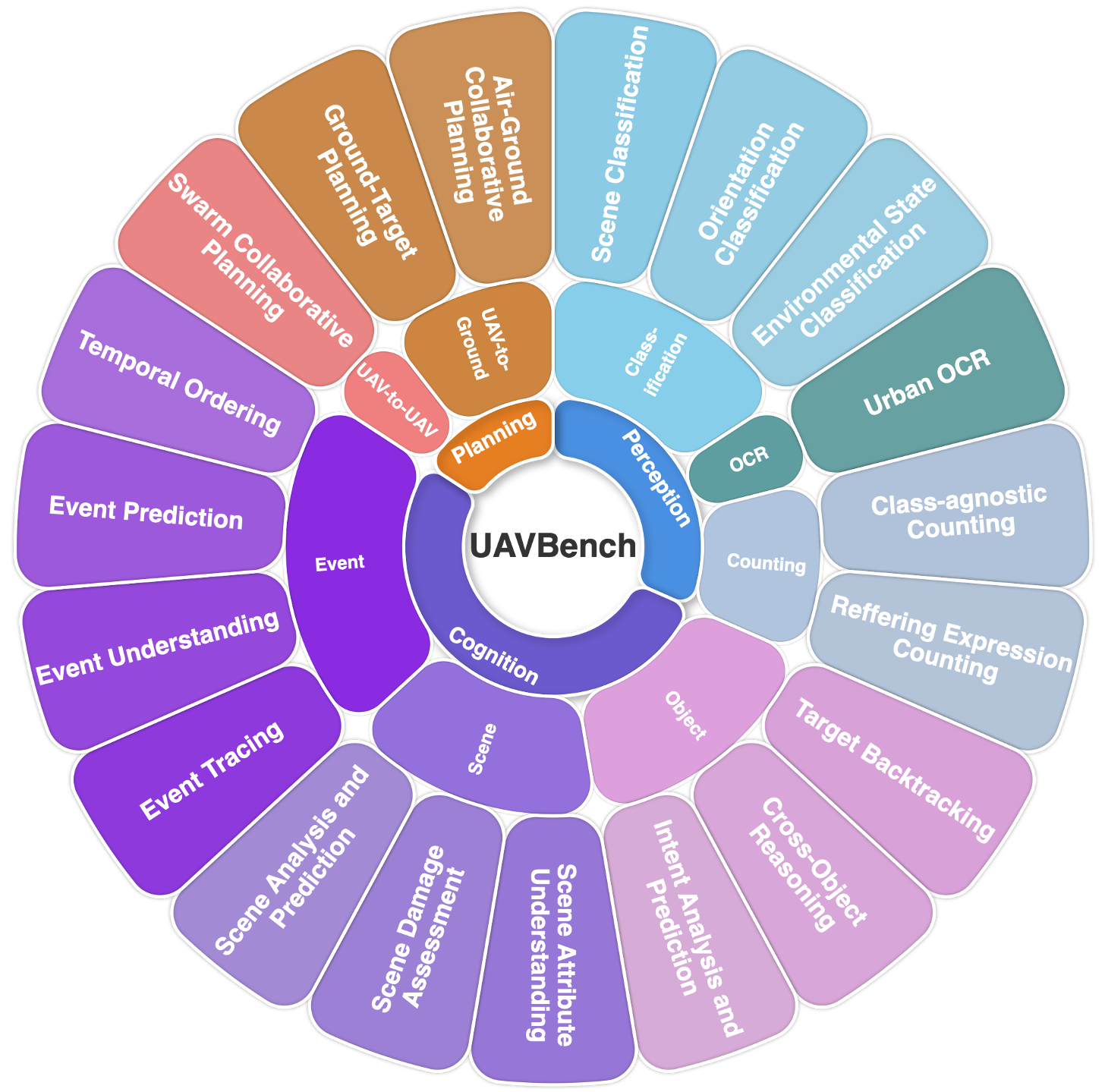}
    \caption{The task design of \OURS covers 3 high-level categories, 8 sub-catigories and 19 fine-grained tasks in \OURS.}
    \label{fig:task}
\end{figure}
In this section, we first introduce the hierarchical task design of \OURS. It spans 3 L1 catigories, 8 sub-catigories, and 19 fine-grained tasks that are specifically tailored for UAV scenarios, as shown in Fig~\ref{fig:task}. Next, we describe the data collection process, question annotation procedures, and quality control guidelines, which greatly enhances the dataset’s reliability and difficulty. Finally, we present a statistical overview of \OURS.

\subsection{Hierarchical Task Design}
The diverse tasks in \OURS are carefully designed to comprehensively evaluate the capabilities of MLLMs, with each task associated with both L1 and L2 categories representing different levels of ability. 
% In this work, we focus on MLLMs as analysis and decision-making models, rather than end-to-end action executors. This separation between decision and action has also been observed in many UAV Vision-and-Language Navigation (VLN) works\citehere. Therefore, we categorize the L2-level abilities under the three core capabilities of perception, cognition, and planning. 
The detailed definitions of each task under L2-level are listed in Appendix.

\textbf{Perception.} This dimension consists of three sub-categories: classification, OCR, and counting. \textbf{1) Classification.} Identifying the category of objects or scenes in images. In UAV scenarios, such tasks include recognizing land-cover types (e.g., roads, buildings, farmlands) and transportation vehicles (e.g., cars, ships, airplanes). In particular, we annotate a large number of vehicle orientation classification tasks, which are crucial for road safety monitoring and trajectory prediction. \textbf{2) OCR (Optical Character Recognition).} Recognizing textual and symbolic information in images, mainly focusing on extracting information from road signs, markings, and traffic signals, which can support navigation and traffic management. \textbf{3) Counting.} Estimating the number of objects such as vehicles, people, or animals. In UAV scenarios, counting is valuable for traffic flow analysis, crowd density monitoring, and wild animal protection.

\textbf{Cognition.} Based on the reasoning target, cognition can be categorized into object-level, scene-level, and event-level reasoning. \textbf{1) Object-level.} Reasoning about the positions and behaviors of single or multiple target objects across past, present, and future spatial-temporal sequences, which supports the analysis of object trajectories, behavioral patterns, and anomalies.\textbf{ 2) Scene-level.} This includes three tasks: scene attribute understanding, scene damage assessment (e.g., fire, flood), and scene flow prediction, aimed at understanding the overall environmental state and its dynamic changes. \textbf{3) Event-level.} Reasoning about the causes, content, prediction, and temporal order of events, which helps UAVs identify events and anticipate their trends.

\textbf{Planning.} Based on the planning entities, planning can be categorized into two types: UAV-to-UAV planning and UAV-to-ground planning(including collaborative) planning.\textbf{1) UAV-to-UAV level.} For a small UAV group (e.g., three UAVs) executing joint missions, planning is conducted from two perspectives: task allocation and fault tolerance. Task allocation is based on the perspective of the UAV with the most comprehensive information (the command UAV), assigning roles and paths to each UAV to optimize overall group efficiency. Fault tolerance ensures that the group can still accomplish its mission even if individual UAVs fail or are disrupted. This capability is critical for tasks such as multi-UAV cooperative tracking, inspection, and search operations. \textbf{2) UAV-to-Ground level.} This level covers Ground-Target Planning and Air–Ground Collaborative Planning, where UAVs guide the movements of ground agents (e.g., rescue teams or vehicles) as well as their own trajectories based on environmental conditions and mission objectives, thereby enabling effective coordination between aerial and ground systems.

\subsection{Dataset Construction}
% UAVBench数据集构建可以分为三部分，收集足够多丰富的UAV场景，进行基于人工和规则转化的两种数据标注流程，设计了很多原则去控制数据集质量，以及最后统计信息展示多样性，
The curation of \OURS can be divided into three main components. 
First, we collect a large and diverse set of real UAV scenarios. 
Second, we adopt two annotation pipelines, manual labeling and rule-based conversion from existing datasets, while following a set of principles to ensure data quality. 
Finally, statistical analysis demonstrates the diversity and comprehensiveness of our benchmark.

\subsubsection{Data Collection}
We collect open-source datasets and conduct re-annotation to construct \OURS. The statistics of annotated datasets for each task are summarized in Appendix. These datasets not only encompass diverse environments such as urban and wilderness settings but also cover extreme scenarios including natural disasters (e.g., floods, wildfires) and human-induced incidents (e.g., violent events, traffic accidents). Moreover, they exhibit substantial diversity across temporal dimensions (day/night, seasonal variations, weather conditions) and geographical dimensions (countries, landscapes). The raw datasets we select follow two main criteria: 1. Data is collected by UAVs in the real world; 2. The datasets contain rich annotation which is beneficial for multiple-choice question generation. For video datasets, we uniformly downsample frames to 12 fps, which both simplifies manual annotation and aligns with mainstream practices in MLLM-based video processing.
\vspace{1.6em}
\begin{figure}[t]
    \centering

    % ---------------- (a) 单栏 ----------------
    \begin{subfigure}{\linewidth}
        \centering
        \includegraphics[width=\linewidth]{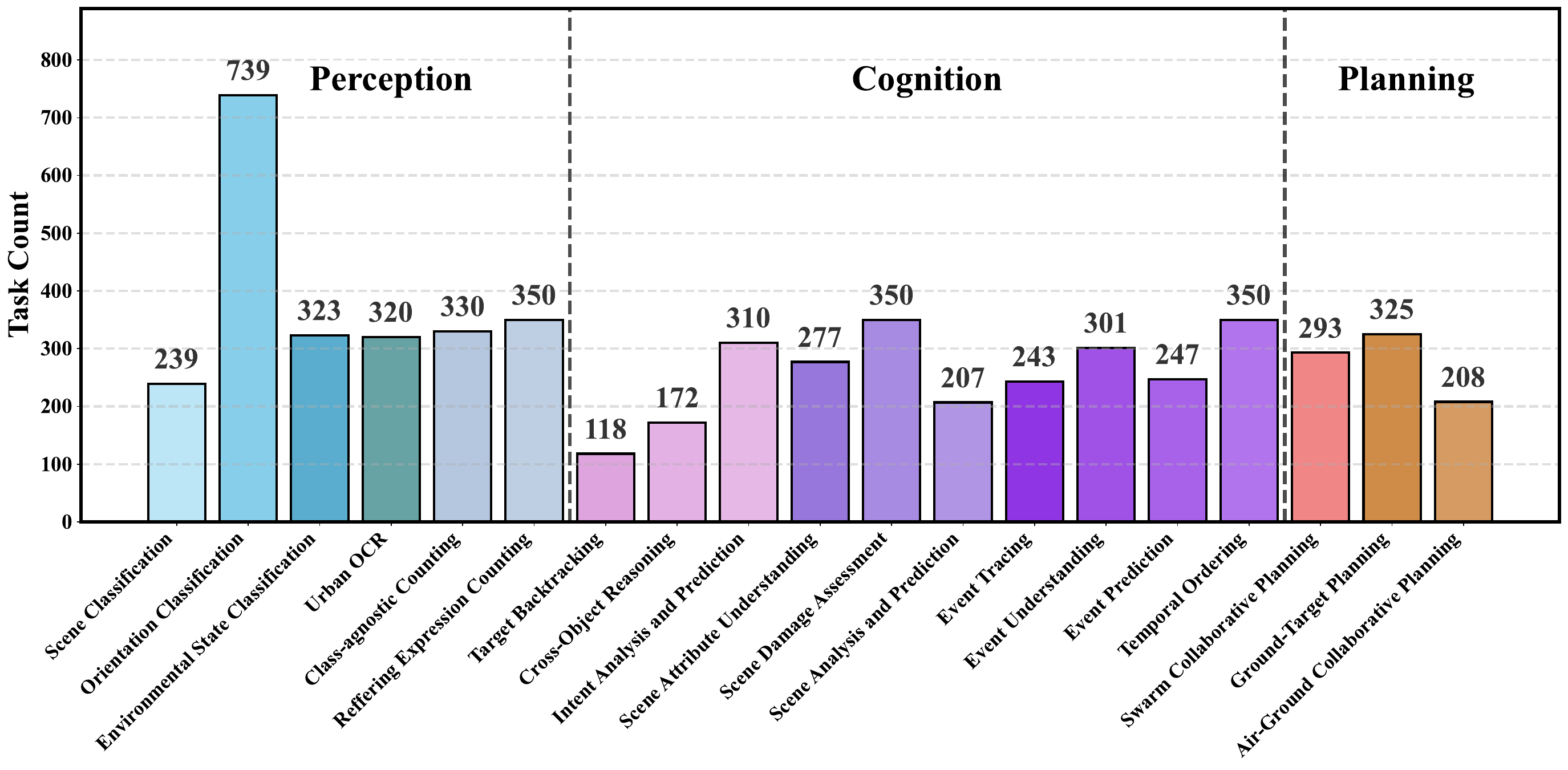}
        \caption{Tasks distribution.}
        \label{fig:stats_a}
    \end{subfigure}

    % ---------------- (b) + (c) 并排顶部对齐 ----------------
    % ----- (b) 饼图 -----
    \begin{minipage}[t]{0.48\linewidth}
        \vspace{0pt}  % 顶端对齐基准点
        \begin{subfigure}{\linewidth}
            \centering
        \includegraphics[width=\linewidth]{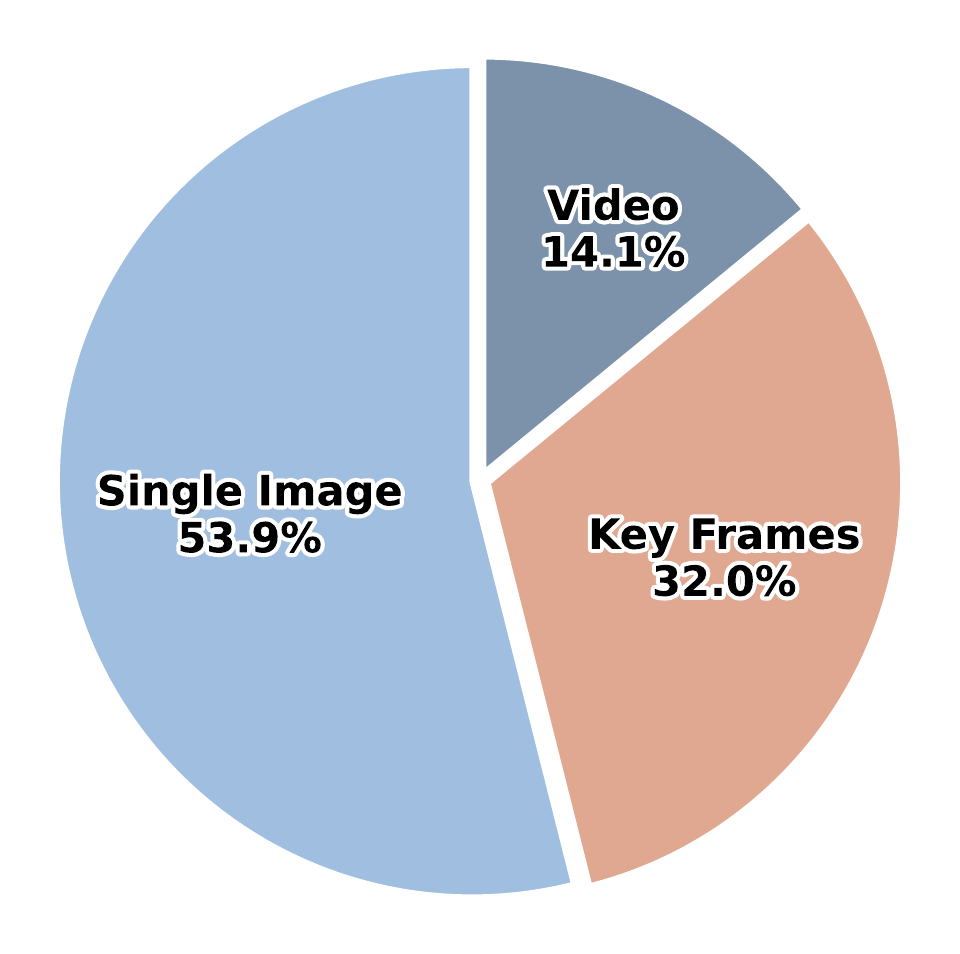}
            \caption{Input modality.}
            \label{fig:stats_b}
        \end{subfigure}
    \end{minipage}
    \hfill
    % ----- (c) 表格 -----
    \begin{minipage}[t]{0.48\linewidth}
        \vspace{0pt}  % 与左侧图保持同一顶端基准
        \begin{subfigure}{\linewidth}
            \centering
            % 表头优化：加粗+调整列宽，内容左对齐增强可读性
           \begin{tabular}{l@{\hspace{0.6em}}c}  % 列间距微调，避免拥挤
                \toprule
                \textbf{Metric} & \textbf{Value} \\  % 用Metric替代Attribute，更贴合量化场景
                \midrule
                $N_{\text{video}}$ & 1549 \\          % 视频数量：N表示数量，下标缩写
                $N_{\text{img}}$ & 2873 \\            % 图像数量
                $\text{Avg. Res.}$ & 1622$\times$1033 \\  % 分辨率保留Avg.缩写
                $N_{\mathrm{bbox,\,man}}$ & 1267 \\
                $N_{\mathrm{bbox,\,obj}}$ & 2560 \\
                $N_{\mathrm{bbox,\,reg}}$ & 3669 \\
                $\text{Avg. } S_{\text{man}}$ & 0.2\% \\ 
                $\text{Avg. } S_{\text{obj}}$ & 0.7\% \\
                $\text{Avg. } S_{\text{reg}}$ & 4.5\% \\
                \bottomrule
            \end{tabular}
            \caption{Annotation metrics.}
            \label{fig:stats_c}
        \end{subfigure}
    \end{minipage}

    \caption{Statistics of \OURS.(a) Distribution of the 19 sub-tasks. (b) Proportions of the three input modalities. (c) Annotation metrics, where $N_{\text{video}}$ and $N_{\text{img}}$ denote the numbers of video clips and images; $\text{Avg.\ Res.}$ denotes the average resolution; $N_{\mathrm{bbox,\,man}}$, $N_{\mathrm{bbox,\,obj}}$, and $N_{\mathrm{bbox,\,reg}}$ denote the numbers of human, object, and region bounding boxes; $\text{Avg.\ }S_{\text{man}}$, $\text{Avg.\ }S_{\text{obj}}$, and $\text{Avg.\ }S_{\text{reg}}$ represent their average area ratios.}
    \label{fig:statistics}
\end{figure}

\subsubsection{Question-Answer Annotation}
We adopt two approaches to construct \OURS: (1) direct human annotation, and (2) rule-based transformation from existing datasets. The detailed construction procedures for all 19 tasks are provided in the Appendix.

\textbf{Human Annotation.}
For most tasks, annotators are provided with predefined task templates and annotate according to appropriate data sources (see Appendix for details). However, for perception-oriented tasks, relying solely on “templates + annotator judgment” is insufficient to control task difficulty. For example, in scene classification, MLLMs may already achieve high accuracy because similar scenes commonly appear in their pretraining corpora.

To address this, we first employ Qwen2.5-VL-72B to synthesize scene classification questions, deliberately increasing task difficulty by enriching the options with fine-grained details. Next, we use Qwen2.5-VL-7B and Qwen2.5-VL-72B answer these synthesized questions and select the cases where the two models disagree  to human annotators. This way raises task difficulty from the data perspective. Furthermore, for tasks where options themselves are complex (e.g., the planning tasks where options may correspond to textual descriptions of distinct routes), we leverage MLLMs to expand the options, enhancing the plausibility of distractors.

\textbf{Transfer from existing datasets.} 
For datasets involving anomaly events and natural disasters, the original annotations usually contain rich semantic and structural information. We first apply rule-based methods to automatically synthesize multiple-choice questions for counting tasks and scene damage assessment. After expert verification, the questions are further refined by MLLMs to fit our task design.

\begingroup
% \definecolor{perceptionColor}{HTML}{87CEEB} 
\definecolor{perceptionColor}{HTML}{96A9BF} 
% \definecolor{cognitionColor}{HTML}{DDA0DD}   
\definecolor{cognitionColor}{HTML}{EBD8B7}
% \definecolor{planningColor}{HTML}{FFA07A}   
\definecolor{planningColor}{HTML}{B5A69C}   
% \definecolor{oai-gray-600}{RGB}{130, 130, 130}
% \definecolor{oai-gray-600}{RGB}{108, 100, 158}
\definecolor{oai-gray-600}{RGB}{232, 159, 44}
% \definecolor{oai-gray-300}{RGB}{220, 220, 220}  
% \definecolor{oai-gray-300}{RGB}{173, 168, 208} 
\definecolor{oai-gray-300}{RGB}{254, 210, 103}
\definecolor{oai-red-1}{RGB}{250, 40, 55}
\definecolor{oai-red-2}{RGB}{255, 110, 120}
\definecolor{oai-red-3}{RGB}{247, 198, 198}

% 导言区需提前加载：\usepackage{graphicx}、\usepackage{xcolor}
\begin{table*}[ht!]  % 自动占满双栏宽度
    \captionsetup{type=table}
    \caption{\textbf{Experimental results on \textbf{\OURS}.} \colorbox{oai-gray-600}{Dark Orange} indicates the best result among all models and \colorbox{oai-gray-300}{light Orange} indicates the best result among open-source models. $\ddag$: We conduct human evaluation on a randomly chosen 10\% subset of the questions from each task.}
    \vspace{-0.3em}
    \centering
    \fontsize{6pt}{7pt}\selectfont  % 增大字体提升可读性
    \setlength\tabcolsep{3pt}  % 拉宽列间距（原2pt→4pt，可按需微调）
    \renewcommand{\arraystretch}{1.5}  % 增大行高，避免文字拥挤
    \begin{tabular}{r|cc|cccccccccccccccccccc}
        & & &
        \rotatebox{75}{Scene. Class.} &
        \rotatebox{75}{Orient. Class.} &
        \rotatebox{75}{Env. State} & 
        \rotatebox{75}{Urban OCR} &
        \rotatebox{75}{CA. Count} &
        \rotatebox{75}{RE. Count} &
        \rotatebox{75}{Target Back.} &
        \rotatebox{75}{Cross-Obj R.} &
        \rotatebox{75}{Intent Anal.} &
        \rotatebox{75}{Scene Attri.} &
        \rotatebox{75}{Scene Damage} &
        \rotatebox{75}{Scene Pred.} &
        \rotatebox{75}{Event Trace} &
        \rotatebox{75}{Event Under.} &
        \rotatebox{75}{Event Pred.} &
        \rotatebox{75}{Temporal Order} &
        \rotatebox{75}{Swarm Plan} &
        \rotatebox{75}{Ground Plan} &
        \rotatebox{75}{Air-Ground Plan} \\
        Methods & Rank & Avg. & 
        \multicolumn{6}{c}{\cellcolor{perceptionColor!50}Perception} & 
        \multicolumn{10}{c}{\cellcolor{cognitionColor!50}Cognition} & 
        \multicolumn{3}{c}{\cellcolor{planningColor!50}Planning} \\
        \hline
        \rowcolor{gray!10}
        \multicolumn{1}{l|}{\textcolor{black}{\textit{Baseline}}} & & & & & & & & & & & & & & & & & & & & & \\
        Random & - & 25.38 & 25.00 & 25.00 & 25.00 & 25.00 & 20.00 & 20.00 & 26.48 & 28.15 & 25.00 & 32.55 & 25.00 & 25.00 & 24.73 & 25.00 & 24.40 & 25.00 & 28.07 & 25.00 & 22.91 \\
        Human{$^\ddag$} & - & 80.39 & 82.61 & 81.94 & 81.62 & 76.67 & 42.42 & 29.41 & 77.50 & 94.12 & 88.72 & 89.78 & 87.80 & 87.50 & 82.61 & 96.67 & 100.00 & 81.78 & 85.71 & 78.26 & 82.35 \\
        \hline
        
        \rowcolor{gray!10}
        \multicolumn{1}{l|}{\textcolor{black}{\textit{API-based}}} & & & & & & & & & & & & & & & & & & & & & \\
        Gemini 2.5 Pro& {\cellcolor{oai-red-1}1} & 54.59 & {\cellcolor{oai-gray-600}74.90} & 37.89 & 73.78 & {\cellcolor{oai-gray-600}82.19} & 23.94 & 24.86 & {\cellcolor{oai-gray-600}51.69} & 48.26 & 50.00 & 84.12 & 44.57 & 57.00 & {\cellcolor{oai-gray-600}73.25} & {\cellcolor{oai-gray-600}73.09} & 68.02 & 51.14 & 25.68 & 44.19 & 48.56 \\
        Gemini 2.5 Flash& {\cellcolor{oai-red-2}2} & 47.44 & 70.71 & 41.00 & 68.30 & 75.94 & 24.55 & {\cellcolor{oai-gray-600}32.86} & 38.98 & 52.91 & 20.97 & 83.39 & 46.86 & 41.55 & 70.78 & 69.44 & 66.40 & 21.14 & 15.54 & 39.94 & 20.19 \\
        % GPT-4o(old) & - & 54.30 & 34.20 & 74.10 & 0.00 & 70.30 & 0.00 & 0.00 & 28.00 & 46.50 & 51.70 & 88.10 & 0.00 & 81.00 & 0.00 & 0.00 & 0.00 & 27.20 & 31.40 & 46.70 & 82.90 \\
        GPT-4o & {\cellcolor{oai-red-3}3} & 44.92 & 57.32 & 38.43 & 68.01 & 62.19 & 22.12 & 18.86 & 14.41 & 25.58 & 50.00 & 83.39 & 31.71 & 40.10 & 67.08 & 64.78 & 63.16 & 33.71 & 28.38 & 39.09 & 45.19 \\
        \hline
    
        \rowcolor{gray!10}
        \multicolumn{1}{l|}{\textcolor{black}{\textit{Open-source}}} & & & & & & & & & & & & & & & & & & & & & \\
        Qwen3-VL-8B & 6 & 50.98 & 69.87 & 26.52 & 71.76 & {\cellcolor{oai-gray-300}79.69} & 38.79 & 18.57 & {\cellcolor{oai-gray-300}50.00} & {\cellcolor{oai-gray-600}57.56} & {\cellcolor{oai-gray-600}58.18} & 85.56 & 34.00 & 53.14 & 58.44 & 62.46 & 61.54 & 30.00 & 27.12 & 45.61 & 39.90 \\
        Qwen3-VL-32B & {\cellcolor{oai-red-1}1} & {\cellcolor{oai-gray-600}55.40} & 68.62 & 41.81 & 75.22 & 76.56 & 45.45 & 26.57 & 42.37 & 54.07 & 50.32 & 87.36 & 40.00 & {\cellcolor{oai-gray-600}64.25} & 64.61 & {\cellcolor{oai-gray-300}69.77} & 71.66 & 42.00 & 37.63 & 50.14 & 44.23 \\
        Qwen3-VL-235B-A22B& {\cellcolor{oai-red-2}2} & 55.07 & 69.04 & 41.14 & {\cellcolor{oai-gray-600}75.79} & 73.44 & 47.27 & 26.00 & 43.22 & 51.74 & 51.94 & 86.28 & 39.43 & 58.45 & 69.96 & 72.43 & 68.42 & 40.57 & 33.45 & 46.74 & {\cellcolor{oai-gray-600}50.96} \\
        Qwen2.5-VL-7B & 7 & 49.64 & 68.62 & 31.12 & 67.72 & 70.31 & 30.00 & 27.71 & {\cellcolor{oai-gray-300}50.00} & 51.16 & 51.94 & 87.73 & 29.71 & 51.21 & 53.91 & 57.81 & 60.73 & 35.43 & 27.12 & 49.01 & 41.83 \\
        Qwen2.5-VL-32B & 5 & 52.02 & 66.95 & 29.63 & 68.88 & 76.56 & 48.79 & 19.14 & 36.44 & 47.67 & 52.90 & {\cellcolor{oai-gray-600}89.53} & 40.29 & {\cellcolor{oai-gray-600}64.25} & 66.67 & 61.13 & 59.11 & 39.43 & 23.73 & {\cellcolor{oai-gray-600}51.56} & 45.67 \\
        Qwen2.5-VL-72B & {\cellcolor{oai-red-3}3} & 54.62 & 70.29 & 34.78 & 71.76 & 75.62 & 35.45 & 24.29 & 38.14 & 50.58 & 57.74 & 89.17 & 25.14 & {\cellcolor{oai-gray-600}64.25} & 71.60 & 66.45 & {\cellcolor{oai-gray-600}72.87} & {\cellcolor{oai-gray-600}56.12} & 35.25 & 50.71 & 47.60 \\
        InternVL3.5-8B & 9 & 47.13 & 58.16 & 27.06 & 72.62 & 68.12 & 36.36 & 17.71 & 34.75 & 48.84 & 36.77 & 88.09 & 34.00 & 49.28 & 54.73 & 56.26 & 60.73 & 34.57 & {\cellcolor{oai-gray-600}40.68} & 39.66 & 37.02 \\
        InternVL3.5-38B & 13 & 43.45 & 46.86 & 22.33 & 55.04 & 66.56 & {\cellcolor{oai-gray-600}49.09} & 18.29 & 27.97 & 29.65 & 47.62 & 63.18 & 24.57 & 52.17 & 67.08 & 59.47 & 63.56 & 35.14 & 32.54 & 24.08 & 40.38 \\
        InternVL3-14B & 10 & 46.86 & 66.53 & 18.40 & 72.05 & 73.75 & 32.42 & 16.29 & 41.53 & 54.07 & 33.87 & 86.28 & 42.00 & 56.52 & 58.44 & 47.84 & 54.25 & 35.71 & 29.49 & 36.26 & 34.62 \\
        InternVL3-78B & 4 & 53.56 & {\cellcolor{oai-gray-300}72.80} & {\cellcolor{oai-gray-600}52.23} & 66.86 & 58.44 & 45.45 & 16.00 & {\cellcolor{oai-gray-300}50.00} & 45.93 & 40.65 & 88.09 & {\cellcolor{oai-gray-600}48.86} & 59.42 & {\cellcolor{oai-gray-600}73.25} & {\cellcolor{oai-gray-300}69.77} & 69.64 & 40.86 & 38.64 & 46.18 & 34.62 \\
        LLaVA-OneVision-7B & 12 & 43.83 & 62.76 & 18.13 & 69.74 & 65.94 & 25.15 & 24.29 & 40.68 & 48.84 & 42.26 & 55.60 & 27.71 & 31.88 & 64.20 & 55.81 & 55.47 & 28.86 & 27.46 & 41.36 & 46.63 \\
        MiniCPM-V-4.5-8B & 8 & 47.70 & 63.18 & 36.27 & 72.05 & 72.50 & 38.79 & 27.14 & 38.14 & 44.77 & 48.06 & 56.68 & 34.57 & 38.16 & 60.49 & 60.80 & 56.28 & 33.14 & 39.32 & 42.78 & 43.27 \\
        MiMo-VL-7B-RL & 11 & 44.33 & 67.36 & 24.22 & 68.88 & 70.62 & 16.97 & {\cellcolor{oai-gray-300}31.43} & 43.22 & 32.56 & 48.06 & 84.48 & 29.14 & 54.11 & 40.74 & 39.20 & 36.84 & 40.00 & 23.73 & 48.44 & 42.31 \\
        \hline
    \end{tabular}
    \label{tab:zero_shot_all}
\end{table*}
\endgroup

\subsubsection{Quality Control}

% UAVBench 的质量控制主要包括两方面：（1）标注准确性控制，确保标注结果的一致性与可靠性；（2）任务难度控制，保证任务对人类和模型均具有合理的挑战性。
The quality control of \OURS consists of two key aspects:
(1) Annotation Accuracy Control, which ensures the consistency and reliability of annotations; and
(2) Task Difficulty Control, which maintains a reasonable challenge level for both humans and models.

% 在标注准确性控制方面，UAVBench 的所有任务均直接或间接来源于人工标注。即使对于从其他数据集中转换的任务，我们也仅选取经过人工标注或人工校验的数据。Each annotation is cross-checked by at least two professional multimodal researchers to ensure accuracy and prevent annotation errors caused by human bias.
% 由于 UAV 场景及任务设计复杂，不同标注者在部分样例上可能难以达成共识。为此，专家团队依据场景语义制定了细化的标注规范。对于事实性任务（即视频中真实发生的事件），我们会明确答题视角，例如在方向或道路描述中，统一说明以无人机或地面目标为中心视角——已有研究指出，大模型往往难以在同一个问题不同视角提问中给出正确回答。
% 对于假设性任务（如规划类任务），问题通常涉及自然灾害或人类活动场景下的路径规划与行为决策，单纯的任务模板很难概括，标注难度较高。我们在设计中通过精选关键帧、增加限定条件，并对场景做合理假设（如骑行者经过受损建筑区域时合理假设其在观察损坏情况）来保证标注的合理性与可复现性。
For annotation accuracy control, all tasks in \OURS originate from human annotations. Even for tasks adapted from existing datasets, we only retain samples that have been manually annotated or verified. Each sample is cross-checked by at least two professional researchers to ensure correctness and reduce annotation bias. Given the complexity of UAV scenes and task instructions, annotators may still disagree on certain samples. To mitigate this, domain experts developed detailed annotation guidelines grounded in scene semantics and the functional roles of UAVs. For factual tasks (i.e., tasks involving events that objectively occur in the video), we further standardize the answering viewpoint. For example, in orientation or road-related descriptions, the annotation protocol explicitly specifies whether the reference frame is the UAV itself or a ground agent. This is essential as prior work shows that MLLMs often produce inconsistent results across different viewpoints of the same question~\cite{sima2025drivelmdrivinggraphvisual}.

% For factual tasks (i.e., events that actually occur in the video), we standardize the answering viewpoint—for example, in orientation or road-related descriptions, we explicitly specify whether the perspective is centered on the UAV or a ground agent. Prior studies have shown that large models often struggle to produce consistent reasoning across different viewpoints of the same question.

% 在任务难度控制方面，我们通过系统的干扰项设计来调节任务挑战度。
% 对于事实性任务，干扰项通常选取与目标在关键帧中共同出现、外观或状态相似的对象，其对应区域面积通常小于图像的 10%，多数情况下甚至低于 1%。
% 对于假设性任务，我们在人工标注过程中引入大模型进行辅助，但其生成的干扰项往往缺乏足够的区分度。针对这一问题，我们在标注与复核阶段重点控制答案与干扰项的粒度，使其聚焦于可明确区分的关键要素，如方向、角度或对象选择（例如 A 与 B 的差异），以保证任务的辨识度和评测有效性。
For task difficulty control, we adjust the challenge level through systematic distractor design.
In factual tasks, distractors are selected from objects that co-occur with the target in keyframes or share similar appearances and states, typically occupying less than 10\%—and in most cases under 1\%—of the image area.
In hypothetical tasks, large models are employed to assist human annotators, but the generated distractors often lack discriminative strength. To mitigate this, we strictly control the granularity of answers and distractors during annotation and review, ensuring they focus on clearly distinguishable factors such as direction, angle, or object choice (e.g., the distinction between Option A and B), thereby maintaining task discriminability and validity.

\subsection{Diversity Statistics}
%QA-Pair、Average Resolution; avg pixels; key frame, bbox
% The statistics information of UAVBench is shown in Figure~\ref{fig:statistics}.
% The proposed datasets include 
% UAVBench最终一共包含19个子任务，共5757个QA pairs，其中82%的任务是来自于人工标注，剩下18%的任务是转换经其他人工标注过的公开数据集。我们的任务覆盖3种输入模态：single image, key frames, video。标注数据覆盖1549个video clips和2894张图片，平均分辨率是1620*1032，即2353281pixels, 最大分辨率可达5472*3648。我们总共标注了7273个bounding box， 分别是region, object, human类型，其bounding box面积占整个输入图片的面积平均为4.7%， 0.7% 和 0.2%。 

The statistical overview of \OURS is presented in Figure~\ref{fig:statistics}. 
\OURS consists of 19 sub-tasks with a total of 5702 QA pairs. 
Among them, 82\% are manually annotated, while the remaining 18\% are converted from publicly available datasets that were originally human-annotated. 
The benchmark covers three input modalities: single images, key frames, and videos, their distribution is shown in Figure~\ref{fig:statistics}b.
In total, the annotated data span 1549 video clips and 2873 images, with an average resolution of 1622$\times$1033. The maximum resolution reaches 5472$\times$3648. 
We overall annotate 7496 bounding boxes across three categories—regions, objects, and humans—whose average areas account for 4.5\%, 0.7\%, and 0.2\% of the corresponding input frame, respectively. Taken together, these statistics demonstrate that \OURS poses a comprehensive and challenging evaluation for MLLMs across diverse tasks, and real-world complexities.

      % or 3_method
\section{Experiment}
\subsection{Settings}
\textbf{Metrics.} 
All questions in \OURS are designed in a multiple-choice format. We report accuracy as the primary evaluation metric. Each model is evaluated three times, and the average accuracy is taken as the final score for each task. For reproducibility, we use a greedy decoding configuration with top\_p = 1.0, temperature = 0.0, and num\_beams = 3.

\textbf{Baselines.}  We select representative proprietary and open-source MLLMs as our baseline models. 
For the proprietary category, we include state-of-the-art models such as GPT-4o~\cite{hurst2024gpt4o}, Gemini 2.5 Pro and Gemini 2.5 Flash~\cite{comanici2025gemini2_5}.
For the open-source category, we adopt Qwen3-VL and Qwen2.5-VL series~\cite{bai2025qwen25vl}, InternVL3.5 and InternVL3 series~\cite{zhu2025internvl3,wang2025internvl3_5}, MiniCPM-V 4.5~\cite{yao2025minicpmv,yu2025minicpmv4_5}, LLaVA-OneVision-7B~\cite{li2024llavaonevision} and MiMo-VL-7B-RL~\cite{coreteam2025mimovl}. More details of the evaluation are provided in Appendix.

\subsection{Quantitative Results}
The quantitative result is shown in Tab.~\ref{tab:zero_shot_all}. We rank the performance of all evaluated MLLMs and highlight the best scores in color. The main conclusions are summarized as follows:

%结论1：无论是general的还是geospatial类型的mllms, 现有的mllms都不能很好的适应无人机场景任务。综合表现最好的模型是xxx，开源中表现最好的模型是xxx。整体上比人类差很多

% 模型内部差异，参数量大小、适配领域
% 模型参数量影响评测分数，但是影响有限
%结论2：模型参数量影响评测分数。可以观察到几乎模型参数量大的模型得分偏高，参数量小的模型得分偏低。这影响面向低空场景的多模态模型的部署方式

% 结论3:维度级别的观点
%结论3（如果测了视频）。评测设置只输入跟问题相关的关键帧的性能总体比输入视频帧的强，说明当前mllms很难把握无人机场景领域的关注重点。

\begin{itemize}[itemsep=7pt]
\item \textbf{Limited adaptability of current MLLMs to UAV scenarios.} While human evaluators achieve 80.4\% average accuracy on our benchmark, existing MLLMs still struggle to adapt effectively to low-alititude UAV tasks. Notably, human performance achieve high scores on cognition and planning tasks, ranging from 78\% to 100\%, however except one task in cognition, the best scores of MLLMs ranging from 40\% to 73\%.
Among all models, Gemini 2.5 Pro achieves the best overall performance, while Qwen3-VL-32B ranks highest among open-source models.

\item \textbf{Proprietary and open-source MLLMs perform comparably.} 
Our results show that the performance gap between proprietary and open-source MLLMs is not significant. For example, the overall performance of Gemini 2.5 Flash and GPT-4o is around the median among open-source models. 
This suggests that the challenges posed by UAV scenarios are a general issue for current MLLMs, regardless of whether they are proprietary or open-source.

\item \textbf{Model size influences performance.} 
We observe a clear trend that models with larger parameter scales tend to achieve higher accuracy, whereas smaller models generally perform worse. 
This finding highlights a potential trade-off between performance and deployability for MLLMs in low-altitude UAV scenarios.

\end{itemize}

\subsection{Influence of Object Scale}
\begin{figure}[t]
    \centering
    \includegraphics[width=1.0\linewidth]{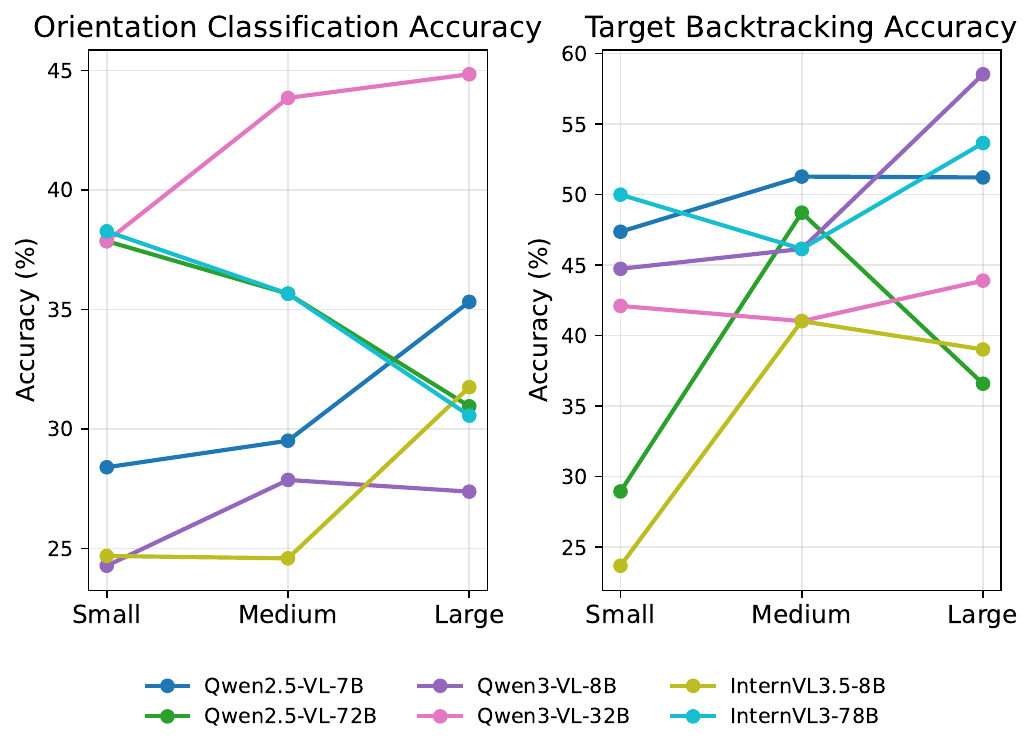}
    \caption{Accuracy comparison across small, medium, and large target sizes on Orient. Classification and Target Backtracking tasks.}
    \label{fig:object_scale}
\end{figure}

% The results in Table \ref{tab:target_scale} indicate that model performance 在不同的任务上受到target scale的影响不同。在感知类任务上和认知类对时序位置理解的任务上，增大目标尺度会模型回答准确率；但是对于考察语义推理能力的任务，对目标尺度依赖较小，表现差异主要来自模型架构与语义训练。

% The results in Table \ref{tab:target_scale} show that the influence of target scale differs across task types. For perception-oriented and spatiotemporal reasoning tasks, larger target sizes generally lead to higher accuracy, although the improvement is not strictly monotonic across all models. This indicates that visual clarity and object salience remain crucial for accurate perception and spatial reasoning. In contrast, semantic reasoning tasks exhibit much weaker dependence on target scale—performance variations in these tasks are mainly attributed to differences in model architecture, multimodal alignment, and the quality of semantic training rather than visual resolution.

We further analyze how object scale affects model performance. For the two tasks with annotated target bounding boxes—Orientation Classification and Target Backtracking—we group all questions into three subsets (Small, Medium, Large) based on the size of the referenced target. We then compute the accuracy for each subset, as summarized in Figure~\ref{fig:object_scale}. Overall, accuracy tends to improve from small to large targets across models, indicating that current MLLMs struggle when the object occupies only a small portion of the field of view. This trend highlights object scale as a key factor shaping model performance in UAV scenarios.

\subsection{Spatial Prediction Bias in MLLMs}
% 从Perception任务Orientation Classification中，模型预测的混淆矩阵见表X。显示出不同模型对于空间状态预测的偏好。比如Qwen2.5VL系列中在Qwen2.5VL72B 几乎不预测车辆向左转还是向右转，Qwen2.5VL32B偏好预测车辆向左转，Qwen2.5-VL-7B-Instruct偏好预测车辆直行和向左转。InternVL3系列模型也表现出来不同的预测偏好。这说明由于数据量、结构差异或对空间关系建模能力不同

\begin{figure}[t!]
    \centering

    % 第一行
    \begin{subfigure}{0.48\linewidth}
        \includegraphics[width=\linewidth]{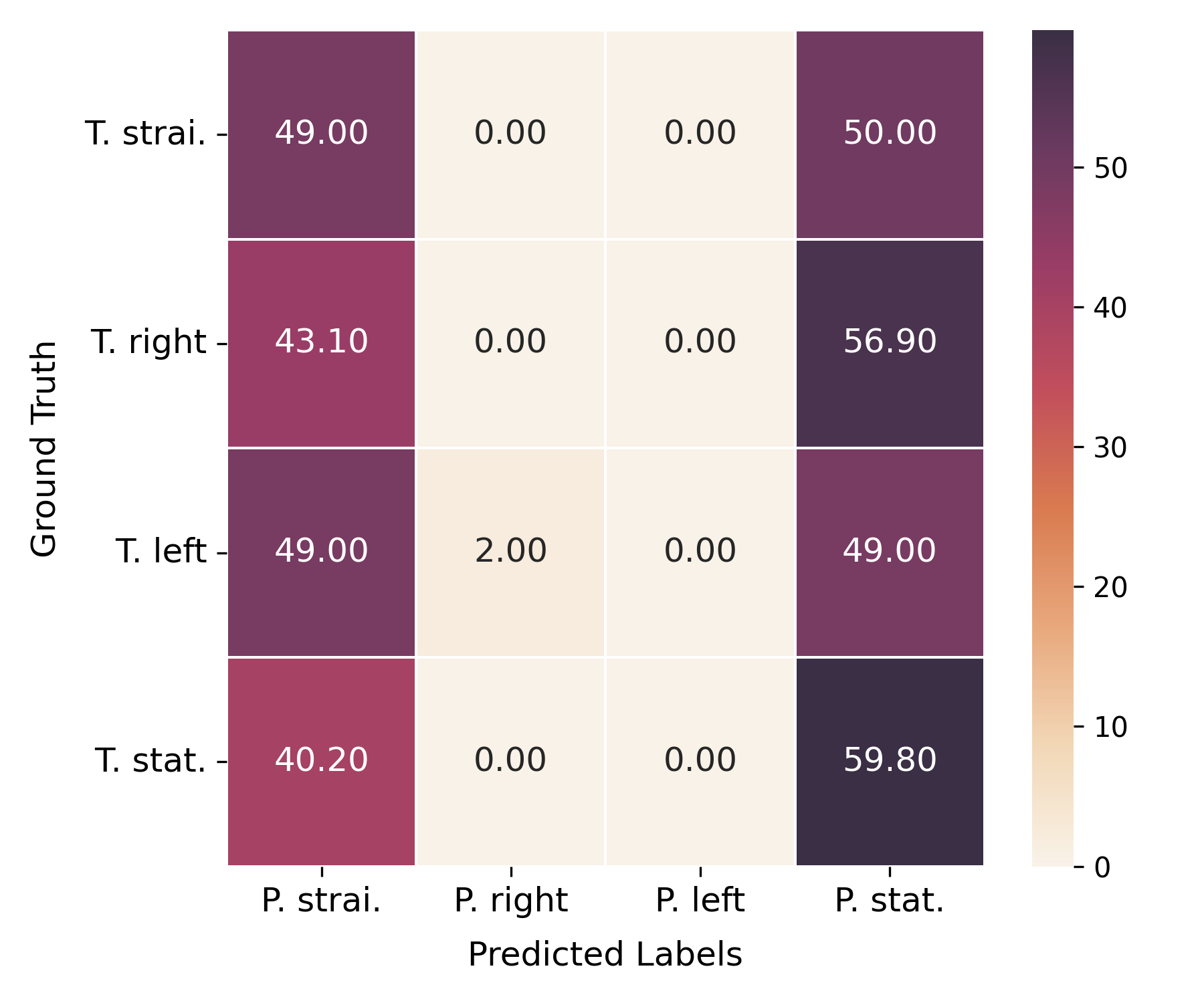}
        \caption*{(a) Qwen2.5VL-72B}
    \end{subfigure}
    \hfill
    \begin{subfigure}{0.48\linewidth}
        \includegraphics[width=\linewidth]{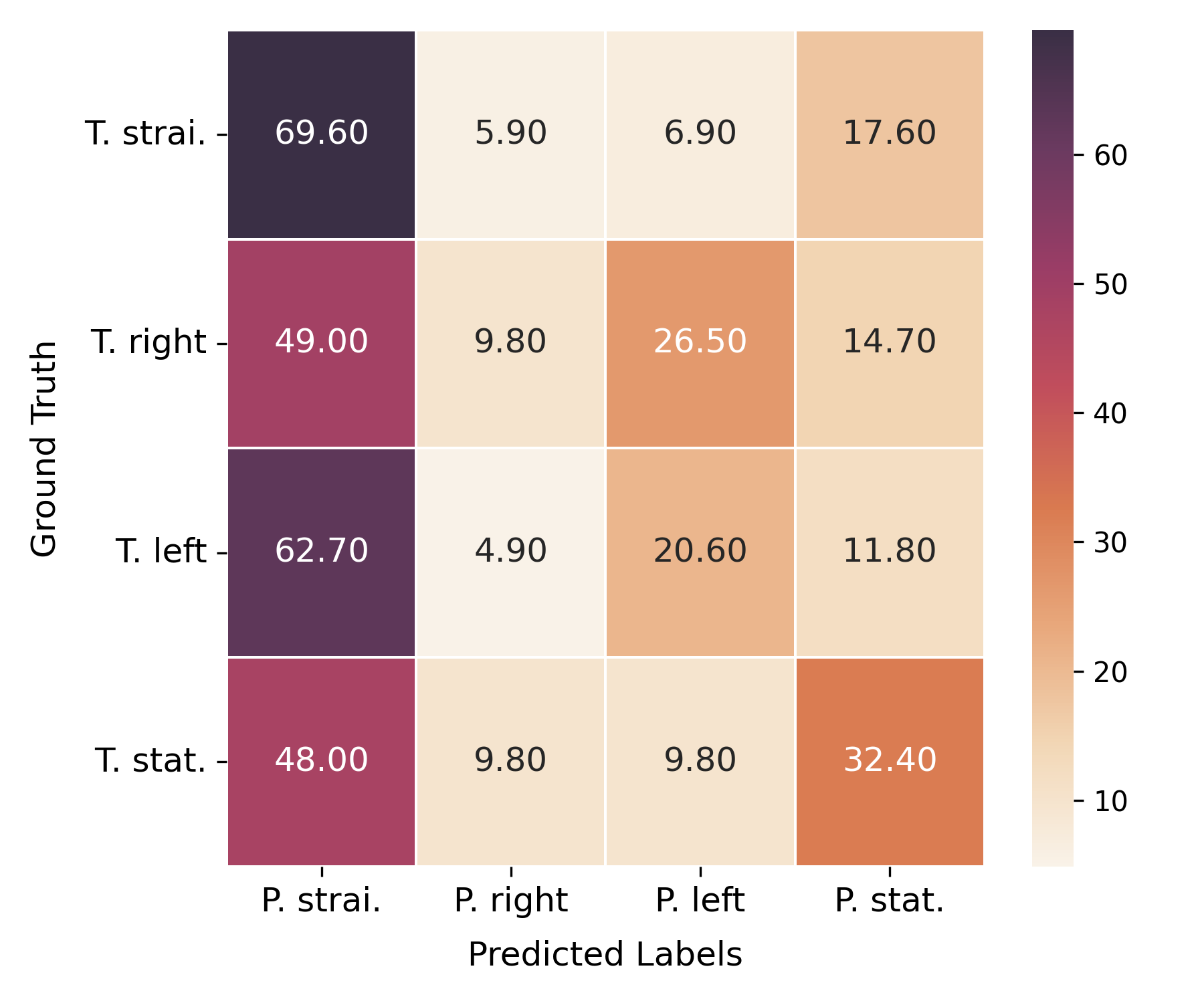}
        \caption*{(d) InternVL3-78B}
    \end{subfigure}

    % 第二行
    \begin{subfigure}{0.48\linewidth}
        \includegraphics[width=\linewidth]{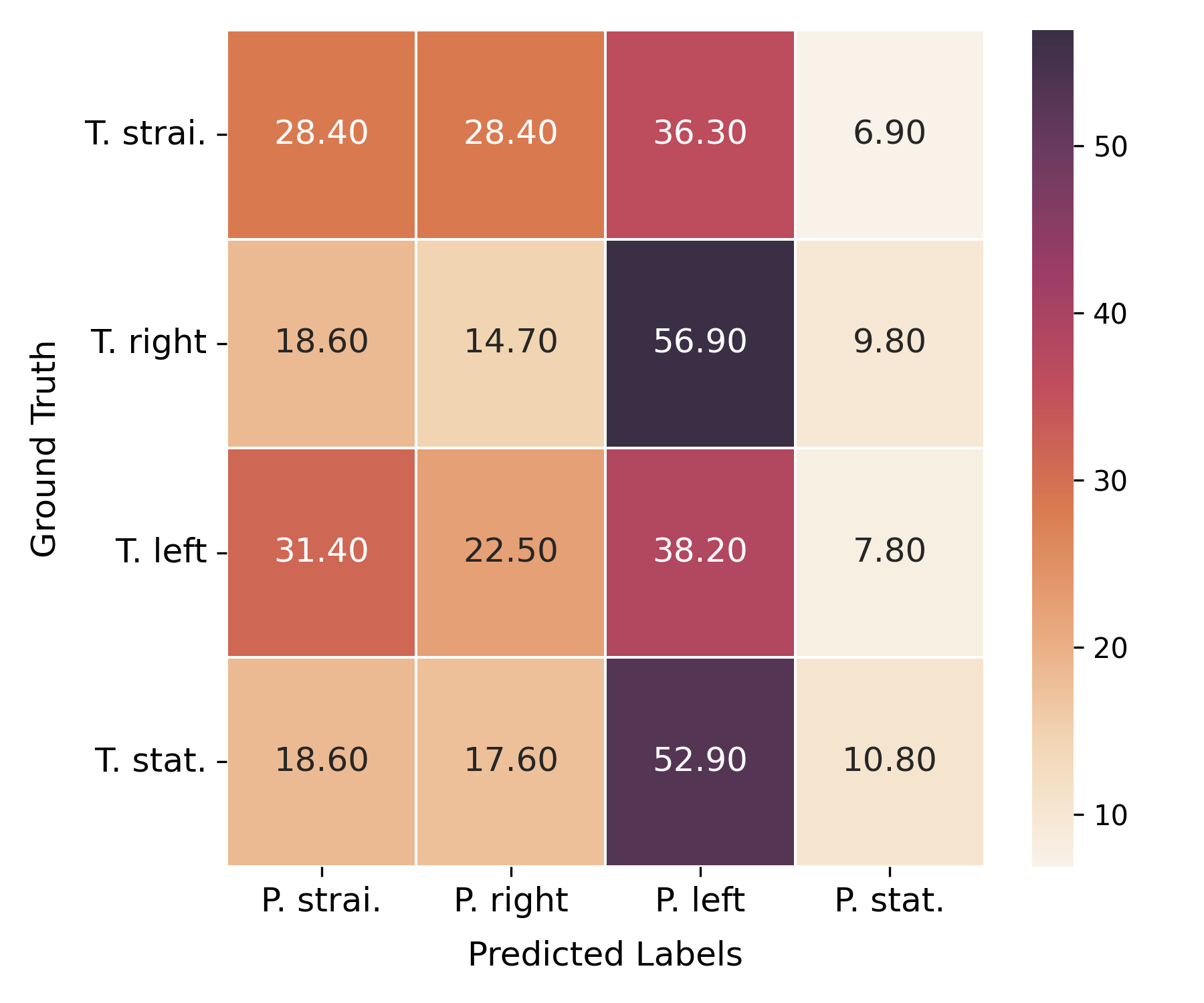}
        \caption*{(b) Qwen2.5VL-32B}
    \end{subfigure}
    \hfill
    \begin{subfigure}{0.48\linewidth}
        \includegraphics[width=\linewidth]{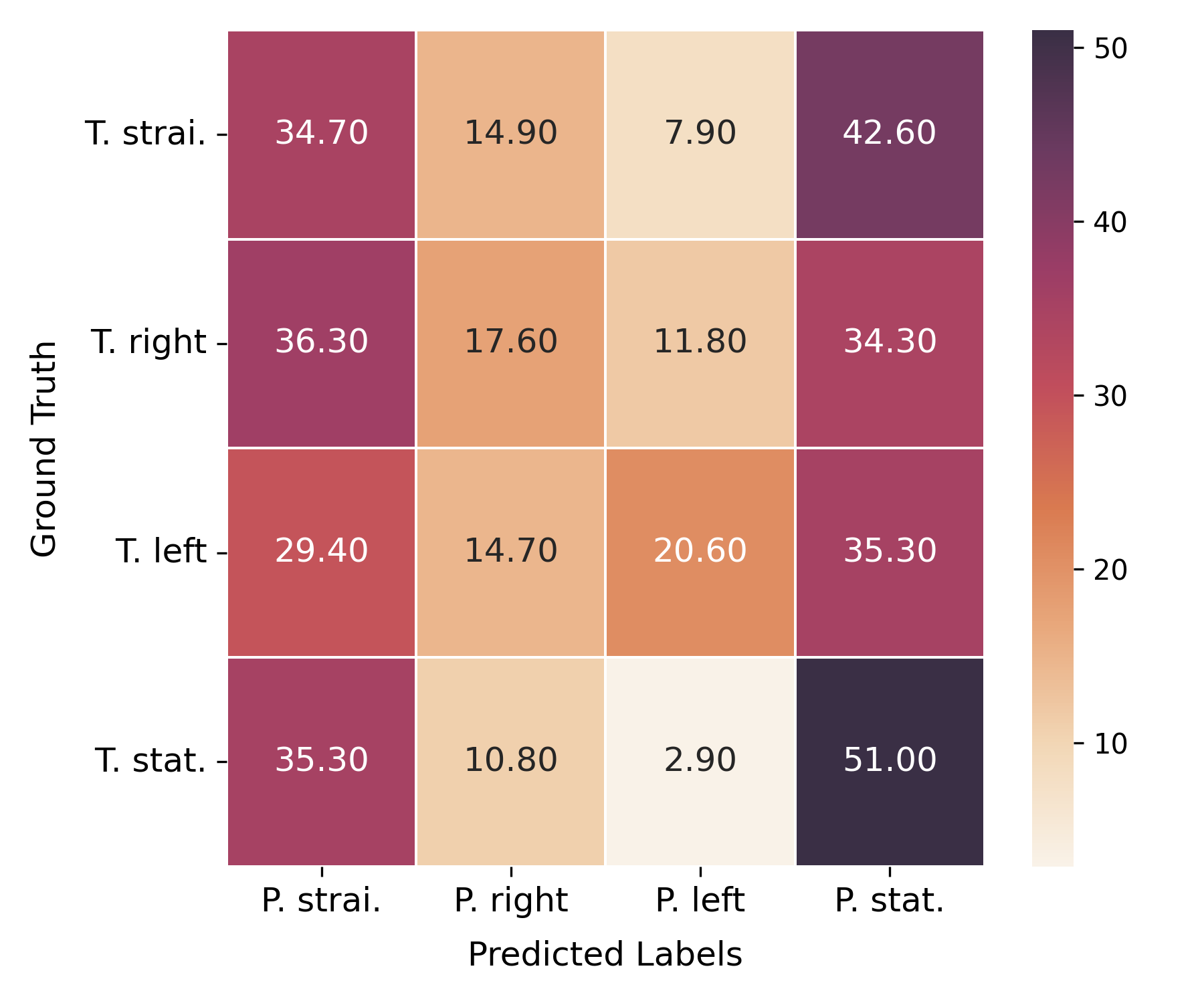}
        \caption*{(e) GPT4o}
    \end{subfigure}

    % 第三行
    \begin{subfigure}{0.48\linewidth}
        \includegraphics[width=\linewidth]{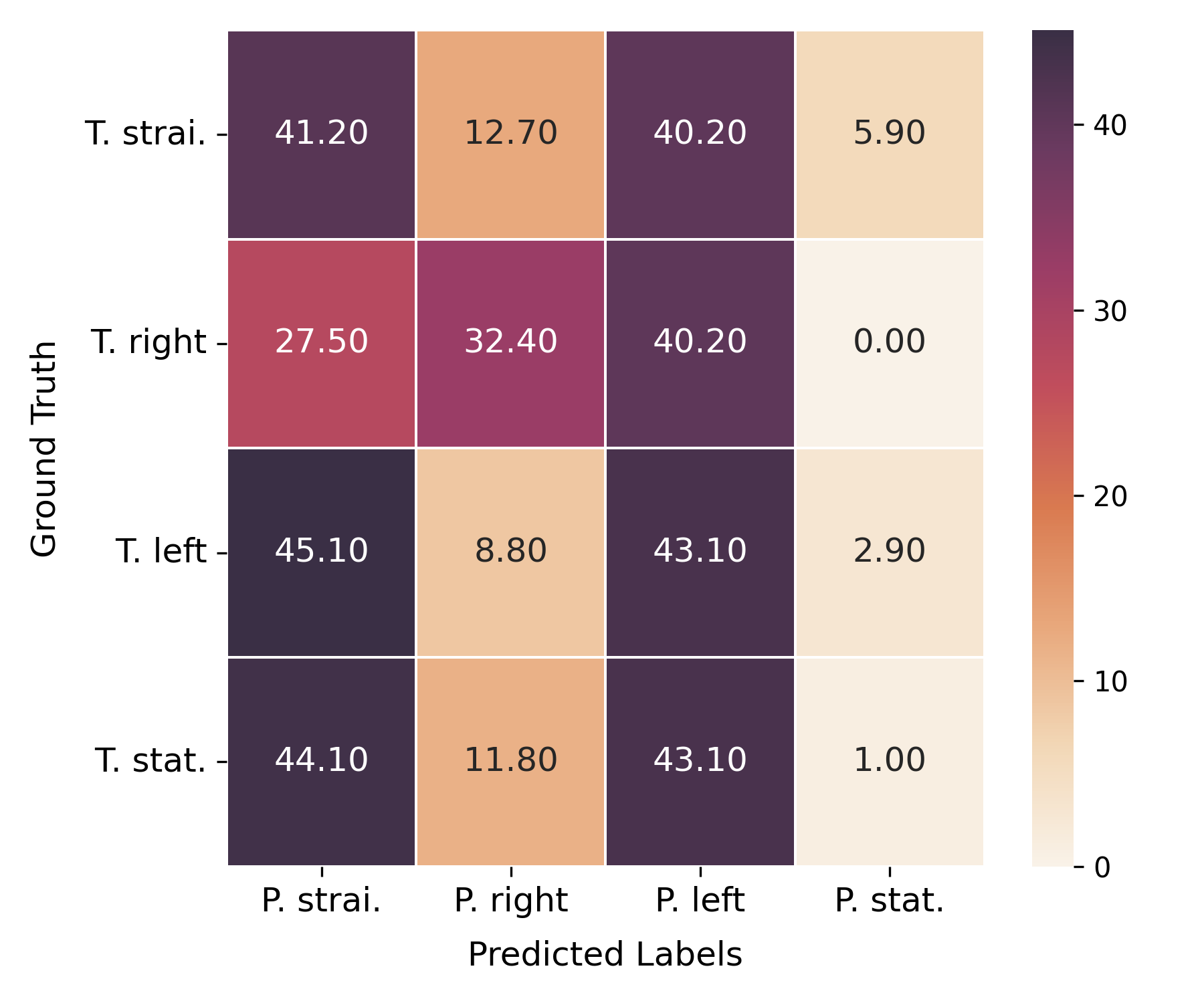}
        \caption*{(c) Qwen2.5VL-7B}
    \end{subfigure}
    \hfill
    \begin{subfigure}{0.48\linewidth}
        \includegraphics[width=\linewidth]{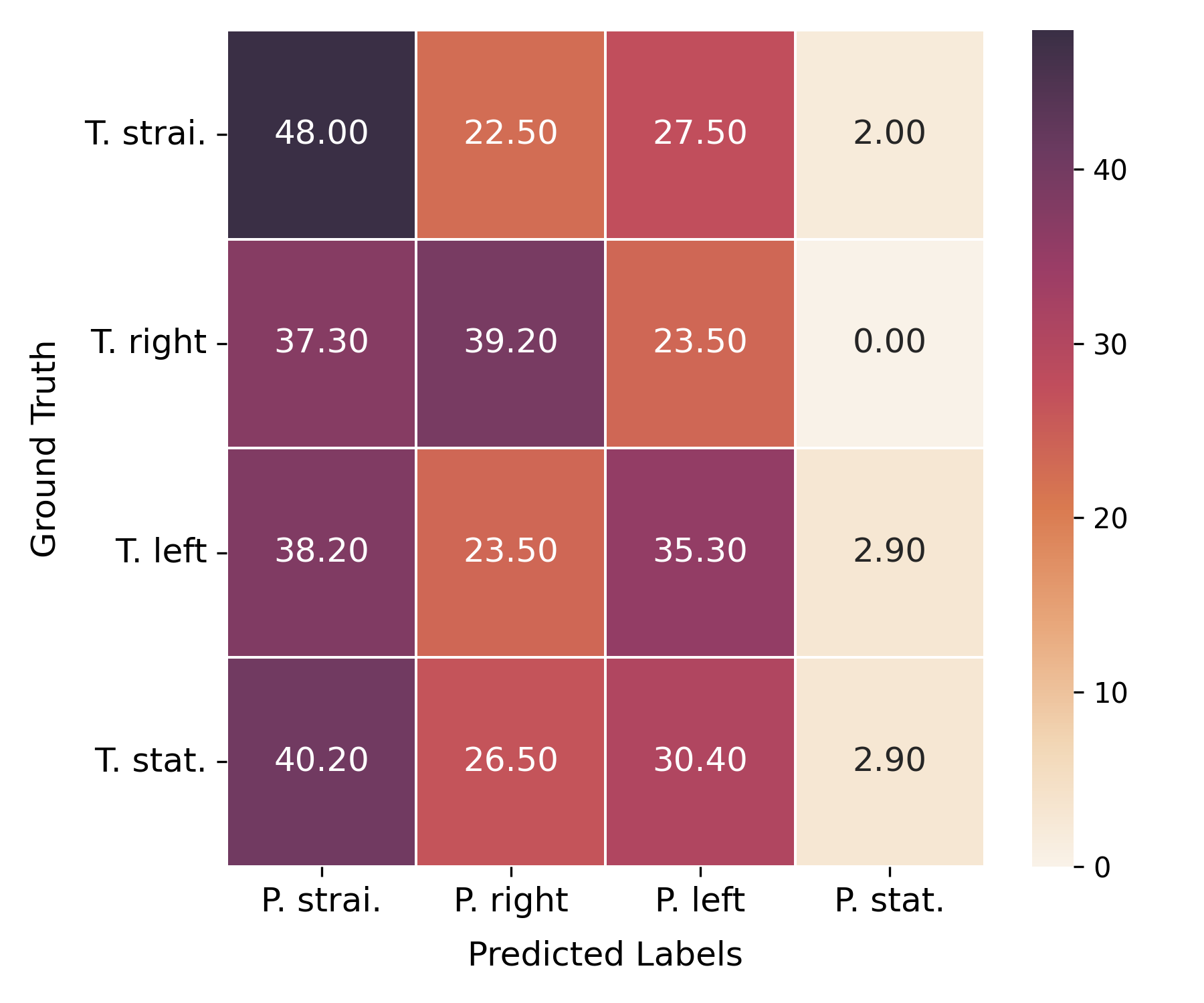}
        \caption*{(f) MiniCPM-V-4.5}
    \end{subfigure}

    \caption{Confusion matrices of predicted (P.) directions from MLLMs versus ground-truth (T.) on the Orient. Classification task.}
    \label{fig:spatial_bias_singlecol}
\end{figure}

We randomly sample 100 instances from each of the four orientation categories in the Orientation Classification task and compare the model predictions with the ground-truth labels. As shown in Fig.~\ref{fig:spatial_bias_singlecol}, the confusion matrices reveal strong model-dependent biases in orientation prediction. Qwen2.5-VL-72B almost never predicts left or right turns, whereas Qwen2.5-VL-32B and Qwen2.5-VL-7B tend to favor turning left and going straight, respectively. Other models exhibit similarly skewed behaviors. This highlights the difficulty current MLLMs face in extracting reliable motion cues from UAV perspectives.

% As shown in Tab.XXX, the confusion matrices obtained from the \textit{Orientation Classification} task under the \textit{Perception} category reveal distinct prediction biases across different models. 
% For instance, within the Qwen2.5-VL series, \textit{Qwen2.5-VL-72B} rarely predicts whether vehicles are turning left or right, while \textit{Qwen2.5-VL-32B} exhibits a tendency to predict turning left and \textit{Qwen2.5-VL-7B-Instruct} tends to predict both going straight and turning left.
% Similarly, the \textit{InternVL3} series also demonstrates varying directional biases across its variants. 
% These differences may stem from the models’ varying capacities in spatial reasoning and motion understanding when interpreting UAV images.

\subsection{Challenges in Multi-View Understanding}
\captionsetup[table]{font=small}

\sisetup{
  round-mode = places,
  round-precision = 2,
  table-number-alignment = center,
  detect-weight = true,
  input-decimal-markers = {.},
}

% \begin{table}[t]  % 单栏表格用table环境
% \centering
% \caption{Accuracy of different MLLMs under single-view (Aerial, Ground) and multi-view settings, grouped into Aerial & Ground Views (left) and Aerial Multi-Views (right).}
% \scriptsize  % 进一步缩小字号适配单栏
% \setlength{\tabcolsep}{5.5pt}  % 大幅减列距（原5.5pt→2.5pt）
% \renewcommand{\arraystretch}{1.5}

% \begin{tabular}{l *{6}{S}}  % 6列数字，无竖线更紧凑
% \toprule
% \rowcolor{gray!15} \textbf{Model} &
% \multicolumn{3}{c}{\textbf{Aerial \& Ground Views}} &  % 表头缩写
% \multicolumn{3}{c}{\textbf{Aerial Multi-Views}} \\  % 表头缩写
% \cmidrule(lr){2-4}\cmidrule(l){5-7}
% & {Aerial} & {Ground} & {Both}
% & {View1} & {View2} & {Both} \\
% \midrule
% GPT-4o    & 44.12 & 61.76 & 57.35 & 45.29 & 50.36 & 49.09 \\
% Qwen3-VL-235B & 52.94 & 55.88 & 47.06 & 54.35 & 55.07 & 52.54 \\  % 模型名缩写
% Qwen3-VL-32B & 67.65 & 52.94 & 61.76 & 51.09 & 48.91 & 48.91 \\
% Qwen2.5-VL-7B & 61.76 & 64.71 & 70.59 & 48.91 & 48.91 & 49.64 \\
% Qwen2.5-VL-72B & 76.47 & 79.41 & 55.88 & 35.14 & 35.51 & 57.97 \\
% InternVL3.5-8B & 44.12 & 38.24 & 38.24 & 36.59 & 36.96 & 36.59 \\
% InternVL3.5-38B & 35.29 & 32.35 & 35.29 & 42.03 & 45.29 & 42.75 \\
% InternVL3-78B & 44.12 & 41.18 & 44.12 & 44.57 & 41.67 & 40.22 \\
% MiniCPM-V-4.5 & 52.94 & 61.76 & 52.94 & 54.35 & 51.81 & 51.45 \\  % 模型名缩写
% MiMo-VL-7B-RL & 64.71 & 64.71 & 55.88 & 47.83 & 49.28 & 47.10 \\
% \bottomrule
% \end{tabular}
% \label{tab:view_settings}
% \end{table}

\begin{table}[t]
\centering
\caption{Accuracy of different MLLMs under single-view (Aerial, Ground) and multi-view settings on the Intent Analysis and Prediction task. 
We additionally report the performance gap between multi-view and the best single view in each group ($\Delta$).}
\scriptsize
\setlength{\tabcolsep}{1.5pt}
\renewcommand{\arraystretch}{1.4}

\begin{tabular}{l *{3}{S} c *{3}{S} c}
\toprule
\rowcolor{gray!15}
\textbf{Model} &
\multicolumn{4}{c}{\textbf{Aerial \& Ground Views}} &
\multicolumn{4}{c}{\textbf{Aerial Multi-Views}} \\
\cmidrule(lr){2-5}\cmidrule(l){6-9}
& {Aerial} & {Ground} & {Both} & {$\Delta$} 
& {View1} & {View2} & {Both} & {$\Delta$} \\
\midrule
GPT-4o          & 44.12 & 61.76 & 57.35 & \negcolor{-4.41}
                & 45.29 & 50.36 & 49.09 & \negcolor{-1.27} \\

Qwen3-VL-235B   & 52.94 & 55.88 & 47.06 & \negcolor{-8.82}
                & 54.35 & 55.07 & 52.54 & \negcolor{-2.53} \\

Qwen3-VL-32B    & 67.65 & 52.94 & 61.76 & \negcolor{-5.89}
                & 51.09 & 48.91 & 48.91 & \negcolor{-2.18} \\

Qwen2.5-VL-7B   & 61.76 & 64.71 & 70.59 & \poscolor{+5.88}
                & 48.91 & 48.91 & 49.64 & \poscolor{+0.73} \\

Qwen2.5-VL-72B  & 76.47 & 79.41 & 55.88 & \negcolor{-23.53}
                & 35.14 & 35.51 & 57.97 & \poscolor{+22.46} \\

InternVL3.5-8B  & 44.12 & 38.24 & 38.24 & \negcolor{-5.88}
                & 36.59 & 36.96 & 36.59 & \negcolor{-0.37} \\

InternVL3.5-38B & 35.29 & 32.35 & 35.29 & 0.00
                & 42.03 & 45.29 & 42.75 & \negcolor{-2.54} \\

InternVL3-78B   & 44.12 & 41.18 & 44.12 & 0.00
                & 44.57 & 41.67 & 40.22 & \negcolor{-4.35} \\

MiniCPM-V-4.5   & 52.94 & 61.76 & 52.94 & \negcolor{-8.82}
                & 54.35 & 51.81 & 51.45 & \negcolor{-2.90} \\

MiMo-VL-7B-RL   & 64.71 & 64.71 & 55.88 & \negcolor{-8.83}
                & 47.83 & 49.28 & 47.10 & \negcolor{-2.18} \\
\bottomrule
\end{tabular}
\label{tab:view_settings}
\end{table}

% 在意图分析与预测任务中，我们进一步研究多模态大语言模型（MLLMs）能否对多视角信息进行推理。如表XXX所示，当前模型的多视角准确率处于两种单视角准确率之间，而人类在整合多个视角时能取得显著更高的性能。这揭示了一种“1 + 1 < 2”现象，表明当前的多模态大语言模型无法实现互补视觉输入的协同整合。这些模型没有利用跨视角的上下文线索来增强推理，而是倾向于独立处理每个视角，这说明它们缺乏有效的跨视角表征对齐和信息融合能力。这些发现凸显了在未来多模态大语言模型的设计中，需要改进多视角组合推理和跨视角注意力机制。

We further decompose the multi-view Intent Analysis and Prediction task into its single-view counterparts to assess whether current MLLMs can effectively leverage complementary cross-view information. Specifically, for each multi-view sample, we evaluate the model independently on each available view to obtain its single-view performance. As shown in Table~\ref{tab:view_settings}, multi-view performance for most models does not surpass the best single-view result. The $\Delta$ column clearly indicates that, in both the Aerial–Ground and Aerial Multi-View settings, multi-view accuracy is frequently lower than that of the strongest single view—revealing a distinct “$1+1<2$” effect. Only a few models (e.g., Qwen2.5-VL-7B and Qwen2.5-VL-72B in the aerial multi-view case) achieve positive gains, while the majority show negative or marginal improvements. These findings demonstrate that current MLLMs lack effective view fusion and fail to combine complementary perspectives into stronger predictions.

% We further split the multi-view tasks in cognition category into single view tasks and test its capability of use multi-view information.
% As shown in Table~\ref{tab:view_settings}, the multi-view accuracy of current models lies between the two single-view accuracies, whereas humans achieve notably higher performance when integrating multiple views.
% This reveals a ``$1+1<2$" phenomenon, indicating that current MLLMs fail to achieve synergistic integration of complementary visual inputs.
% Instead of leveraging cross-view contextual cues to enhance reasoning, the models tend to process each view independently, suggesting a lack of effective cross-view representation alignment and information fusion.
% These findings highlight the need for improved multi-view compositional reasoning and cross-perspective attention mechanisms in future MLLMs designs.

\subsection{Difficulty in Egocentric Planning}
% \begin{table}[t]
% \centering
% \small
% \setlength{\tabcolsep}{4pt}
% \renewcommand{\arraystretch}{1.1}
% \sisetup{
%   table-format=1.3,  % 1 位整数 + 3 位小数
%   detect-weight=true,
%   detect-inline-weight=math,
%   input-decimal-markers=.,
% }
% \begin{tabular}{l|*{3}{S}}
% \toprule
% \textbf{Model} & \multicolumn{3}{c}{\textbf{Accuracy}} \\
% \cmidrule(lr){2-4}
% & \textbf{Subject plan.} & \textbf{Object plan.} & \textbf{Mixed} \\
% \midrule
% Qwen3-VL-8B & 50.00 & 59.18 & 0  \\
% Qwen3-VL-32B &  60.20 & 65.31 & 0  \\
% Qwen2.5-VL-72B & 56.12 & 58.16 & 0  \\
% InternVL3.5-8B & 46.94 & 52.41 & 0  \\
% InternVL3.5-38B & 44.90 & 47.96 & 0  \\
% InternVL3-78B & 54.08 & 51.02 & 0  \\
% MiMo-VL-7B-RL & 54.08 & 60.20 & 0 \\
% \bottomrule
% \end{tabular}
% \caption{Subject--Object Analysis 1: Subject Planning vs. Object Planning}
% \label{tab:subject_object_analysis_1}
% \end{table}

\captionsetup[table]{font=small}

\sisetup{
  round-mode = places,
  round-precision = 2,
  table-number-alignment = center,
  detect-weight = true,
  input-decimal-markers = {.},
}

\begin{table}[t]
\centering
\caption{Performance comparison between egocentric and exocentric planning, obtained by decomposing the original Air–Ground Collaborative Planning task~(Mixed).}
\scriptsize  % 统一使用scriptsize字号
\setlength{\tabcolsep}{5.5pt}  % 列间距与参考表一致
\renewcommand{\arraystretch}{1.5}  % 行高与参考表一致（1.5）

\begin{tabular}{l *{3}{S}}  % 无竖线设计，与参考表统一
\toprule
\rowcolor{gray!15}  % 表头行浅灰背景，与参考表一致
\textbf{Model} & \textbf{Egocentric Plan.} & \textbf{Exocentric Plan.} & \textbf{Mixed} \\
\midrule
Qwen3-VL-8B        & 50.00 & \textbf{59.18} & 34.44 \\
Qwen3-VL-32B       & 60.20 & \textbf{65.31} & 36.67  \\
Qwen2.5-VL-72B     & 56.12 & \textbf{58.16} & 30.00  \\
InternVL3.5-8B     & 46.94 & \textbf{52.41} & 30.00  \\
InternVL3.5-38B    & 44.90 & \textbf{47.96} & 38.89  \\
InternVL3-78B      & \textbf{54.08} & 51.02 & 28.89  \\
MiMo-VL-7B-RL      & 54.08 & \textbf{60.20} & 26.67 \\
\bottomrule
\end{tabular}
\label{tab:subject_object_analysis_1}
\end{table}
%Situational-Responsive Collaborative Planning任务要求，无人机既要对地面上的目标对象规划，也对自主运动进行规划。我们将该任务拆分成对地面目标规划（exocentric planning)和对自主行动规划（egocentric planning）两部分单独评测，结果见表XXX。可以看到整体上模型对oexocentric的planning能力大于egocentric planning能力。这个发现说明...
% The Air-Ground Collaborative Planning task requires UAVs to plan both for ground targets and for their own self-motion.
% We decompose this task into two subcomponents: exocentric planning, which focuses on the planning of ground objects or other agents, and egocentric planning, which concerns the UAV’s self-oriented decision-making.
% As shown in Table~\ref{tab:subject_object_analysis_1}, models generally exhibit stronger exocentric planning ability than egocentric planning ability. Performance on the Mixed setting, which requires joint reasoning over both ego- and exocentric factors, is substantially lower, indicating difficulty in integrating self- and other-centric cues. 

The Air–Ground Collaborative Planning task requires UAVs to simultaneously plan for ground agents and for their own self-motion. To better examine model behavior, we decompose this task into two subcomponents: exocentric planning, which focuses on predicting or planning for ground objects or other agents, and egocentric planning, which concerns the UAV’s own goal-directed decision-making. As shown in Table~\ref{tab:subject_object_analysis_1}, models consistently perform better on exocentric than on egocentric planning, indicating that they are more adept at interpreting external scene dynamics than reasoning about their own actions. Moreover, performance in the Mixed setting, where both self- and other-centric cues must be integrated, is substantially lower across all models, revealing significant difficulty in combining these two forms of reasoning. These results highlight a fundamental gap in current MLLMs that they struggle to ground predictions in their own embodiment and to generate coherent plans.

\subsection{Other Error Analysis}
\begin{figure}[t]
    \centering
    \includegraphics[width=0.9\linewidth]{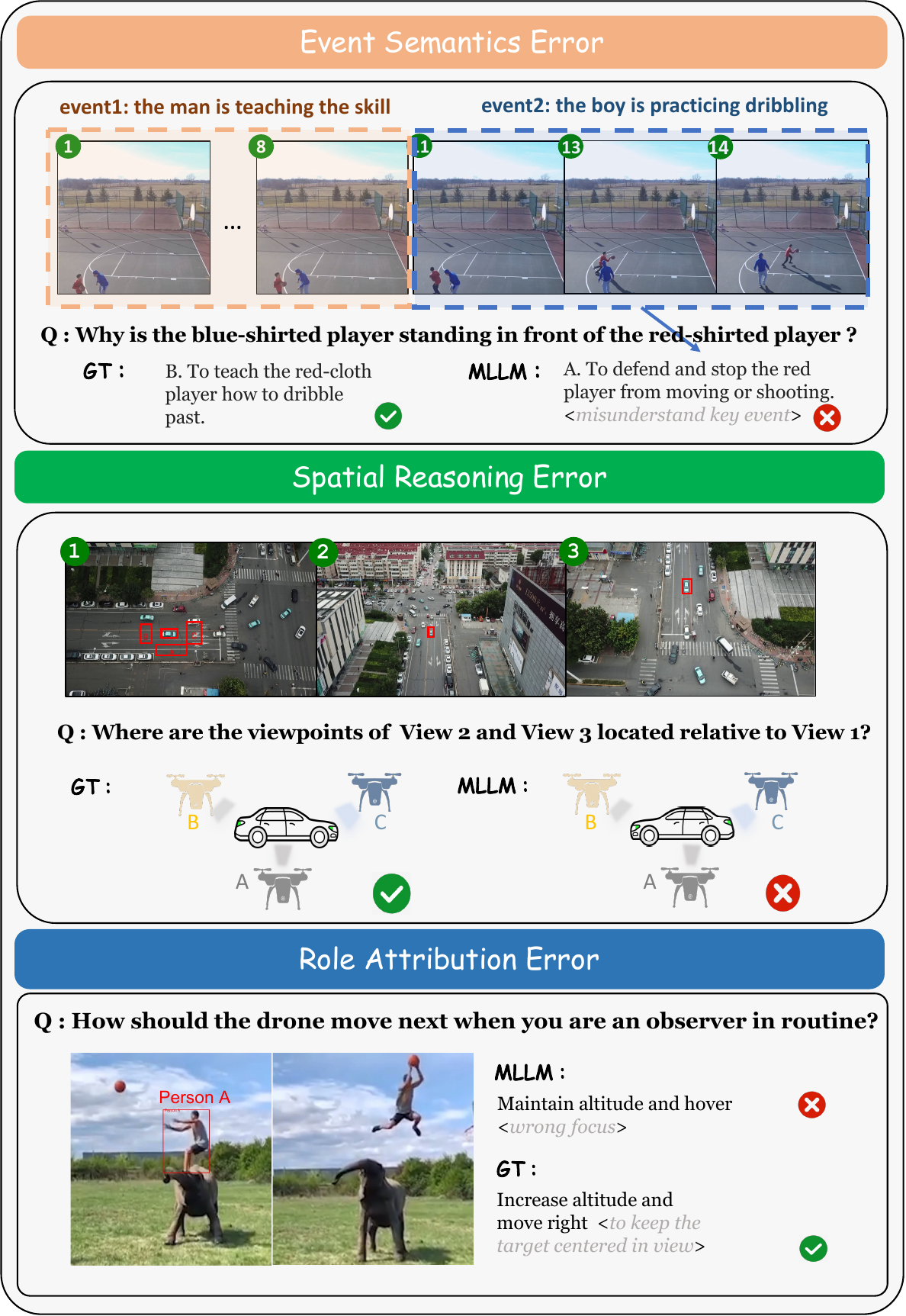}
    % \caption{Qualitative failure analysis of MLLMs on UAVBench: (a) event semantics errors, (b) spatial reasoning errors, and (c) role attribution errors.}
    \caption{Qualitative failure analysis of MLLMs on \OURS.}
    \label{fig:error_analysis}
\end{figure}

We further perform qualitative analysis and identify three additional types of failures beyond the earlier quantitative findings, as illustrated in Figure~\ref{fig:error_analysis}:

\begin{itemize}[leftmargin=1.2em]

    \item \textbf{Event Semantics Error.}
    MLLMs fail to correctly interpret the core semantics of an event—such as who is teaching, practicing, attacking, or defending. Misjudging these key actions leads to incorrect understanding of the event dynamics and flawed subsequent reasoning.

    \item \textbf{Spatial Reasoning Error.}
    MLLMs misinterpret the spatial correspondence between 2D schematic layouts and 3D real-world configurations, resulting in incorrect judgments about UAV viewpoints, relative positions, and coverage relationships across multiple views.

    \item \textbf{Role Attribution Error.}
    MLLMs incorrectly assign semantic roles to the entities involved in the scene—for example, confusing the person who should be tracked or misidentifying who serves as the primary actor. Such role attribution mistakes lead to incorrect predictions and misguided UAV planning decisions.

\end{itemize}

\section{Conclusions}

In this work, we introduce \OURS, a comprehensive benchmark designed to evaluate the perception, cognition, and planning capabilities of multimodal large language models in low-altitude UAV scenarios. \OURS offers a diverse, high-fidelity, and domain-tailored testbed for assessing MLLM performance.
Through extensive evaluations and detailed analyses, we show that while current MLLMs exhibit promising general capabilities, 
% their performance still requires improvement and 
they struggle with UAV-specific challenges such as object-scale variation, spatial perception bias, multi-view understanding, and egocentric planning. These findings highlight a clear gap between generic multimodal intelligence and the requirements of realistic UAV operations.
We hope that \OURS will inspire future research toward more capable, reliable, and UAV-oriented MLLMs for real-world deployment.

% ===== References =====
{\small
\bibliographystyle{ieeenat_fullname}
\bibliography{main}
}

% WARNING: do not include supplementary pages in the main submission
% 附录/补充材料请单独编译为 suppl；若仅本地预览可临时解开：

\clearpage
\setcounter{page}{1}
\maketitlesupplementary

\section{Appendix Outline}
In the supplementary materials, we provide additional details, results, and visualizations to complement the main paper:

\begin{itemize}
    \item \textbf{\OURS Details.} Including L3 sub-task definitions, task templates for annotation, and other annotation details.
    \item \textbf{Evaluation Details.} Including full experimental setup, evaluation prompts, and post-processing after evaluation.
    \item \textbf{Extended Results and Analysis.} Including results on L2 category, results with CoT, more analysis on spacial prediction bias.
    \item \textbf{Visualizations and Challenging Cases.} Including challenging examples from each \OURS task and the corresponding MLLM responses.
\end{itemize}

% Part 1
\section{\OURS Details}
\subsection{Definition of Each Task}
\begin{table*}[t]
\centering
\caption{Task taxonomy of \OURS, including hierarchical categories, definitions, and examples.}
\footnotesize % 保持字体大小适中
\setlength{\tabcolsep}{4pt} % 调整列间距以更好地利用空间
% 使用 array 宏包的 \arraybackslash 确保 p 列中的文本正常换行
% 使用 @{} 移除两边的额外空间

\begin{tabular}{@{}p{1.5cm}|p{1.5cm}|p{2cm}|p{5.5cm}|p{5.5cm}@{}} % 增加垂直线
\toprule
\textbf{L1 Category} & \textbf{L2 Category} & \textbf{L3 Sub-task} & \textbf{Task Definition} & \textbf{Example} \\
\midrule
% ----- Perception -----
\multirow{6}{*}{\textbf{Perception}} & \multirow{3}{*}{Classification} & Scene Classification & Capture and categorize scenes from the entire image or selected areas. & Observe the image. What is the primary type of scene within the red box area?\\
\cmidrule(r){3-5} % 内部细线

 & & Orientation Classification & Identify the ongoing turning behavior of vehicles/people. & Based on visible cues, what is the immediate motion direction of the white SUV within the red box area relative to its own orientation?\\
\cmidrule(r){3-5} % 内部细线
 & & Environment State Classification & Identify the lighting and weather conditions of the scene in the image. & What are the lighting and weather conditions in this image?\\
\cline{2-5} % L2 类别之间的全线分割
 & OCR & Urban OCR & Recognize the text within the specified bounding box in the image. & What is the exact text on the warning sign at the extreme angle?\\
\cline{2-5} % L2 类别之间的全线分割
 & \multirow{2}{*}{Counting} & Class-agnostic Counting & Count the objects of the specified category in the image. & Please count the number of the objects in the image that belong to the category: sheep\\
\cmidrule(r){3-5} % 内部细线
 & & Referring Expression Counting & Count the objects that meet the specific descriptions in the image. & Please count the number of the objects or people that match the description: The white vehicles waiting at the traffic light.\\
\midrule
% ----- Cognition -----
\multirow{10}{*}{\textbf{Cognition}} & \multirow{3}{1.6cm}{Object-Level Reasoning} & Target Backtracking & Trace back the spatial positions and behaviors of the target in the past spatiotemporal sequence. & What did the car do before reaching this point?\\
\cmidrule(r){3-5}
 & & Cross-Object Reasoning & Analyze the behavioral relationships or spatial connections among multiple subjects at the current moment. & Can another car be parked between these two cars?\\
\cmidrule(r){3-5}
 & & Intent Analysis and Prediction & Predict the target's future spatial positions and behaviors based on its current spatial position and behavior. & How will the car in the bounding box travel at the T-junction based on ground and aerial perspectives?\\
\cline{2-5} % L2 类别之间的全线分割
 & \multirow{3}{1.6cm}{Scene-Level Reasoning} & Scene Attribute Understanding & Analyze the attributes or functions of the entire image or the selected region. & How can you describe this scene, Modern or retro?\\
\cmidrule(r){3-5}
 & & Scene Damage Assessment & Analyze the damage degree of the scene/Compare the damage degrees of multiple scenes. & How to rank the severity of building damage across the three scenes (from mild to severe)?\\
\cmidrule(r){3-5}
 & & Scene Analysis and Prediction & Predict the future change trends based on the changes occurring within the scene over a period of time. & Predict the most likely changes in traffic flow within the scene. \\
\cline{2-5} % L2 类别之间的全线分割
 & \multirow{4}{1.6cm}{Event-Level Reasoning} & Event Tracing & Analyze the causes of the event's occurrence. & What led to the black cars closely follow white cars shown in the video?\\
\cmidrule(r){3-5}
 & & Event Understanding & Understand the events that are occurring in the video. & What specific harvesting operation is depicted in the aerial view of the cornfield?\\
\cmidrule(r){3-5}
 & & Event Prediction & Predict the future development trend of the events in the video. & How will the conflict situation shown in the video develop?\\
\cmidrule(r){3-5}
 & & Temporal Ordering & Analyze the chronological order of multiple key frames in a video. & In which order should these images be arranged to match the actual progression of the event?\\
\midrule
% ----- Planning -----
\multirow{3}{*}{\textbf{Planning}} & UAV-to-UAV Planning & Swarm Collaborative Planning & Based on the information provided by multiple drones from different perspectives, select the optimal drone task allocation strategy. & There are three drones providing main, secondary, and third-perspective views. The drone offering the third perspective has withdrawn due to low battery. To maintain the original tracking plan, determine whether a new drone should be deployed and where it should be added.\\
\cmidrule(r){2-5}
 & \multirow{2}{1.5cm}{UAV-to-Ground Planning} & Ground-Target Planning & The ground target needs to complete a specific task; provide a reasonable action plan or route. & If rescuers need to rescue the injured at ABCDE's point after the earthquake and take the wounded to the Region $\langle A \rangle$ for evacuation, please plan the most suitable rescue route.\\
\cmidrule(r){3-5}
 &  &  Air-Ground Collaborative Planning & Provide a reasonable action plan or route for the ground target and drones to jointly complete a specific task. & A religious activity is in progress. Based on the information boxed in the picture, to ensure that Vehicle B can move forward smoothly, what actions should Vehicle A and the drone take?\\
\bottomrule
\end{tabular}

\label{tab:task_taxonomy}
\end{table*}
The hierarchical categories and definition of each task are shown in Table~\ref{tab:task_taxonomy}. This taxonomy provides a comprehensive coverage of UAV-related MLLM capabilities, ranging from basic perception and scene understanding to complex event reasoning and collaborative planning.

\subsection{Annotation Details}
% \subsubsection{Template and data source of each task}

\begin{table*}[t]
\centering
\caption{Task templates and annotation data sources used for each L3 sub-task.}
\label{tab:template_source}
\small % 适当增大字体，提高表格可读性
\setlength{\tabcolsep}{5pt} % 调整列间距
\renewcommand{\arraystretch}{1.2} % 增加行高，避免文本堆叠

% 使用 @{}p{...} 定义列宽，并使用 | 增加 L3 和 Template 之间的垂直线
\begin{tabular}{@{}p{2.5cm}|p{9.5cm}|p{3cm}@{}}
\toprule
\textbf{L3 Sub-task} & \textbf{Task Template} & \textbf{Data Source} \\
\midrule

% ----- Perception -----
% \multicolumn{3}{@{}l}{\textbf{Perception Tasks}} \\ % L1/L2 分组标题
% \midrule
Scene Classification & Given an image or a selected region within the image, what category does the corresponding scene belong to?& Visdrone-DET\cite{zhu2021detection}\\
\cmidrule(r){1-3}
Orientation Classification & Given an image and the selected object, what is its current turning intention based on its own movement direction?& Visdrone-DET\cite{zhu2021detection} \\
\cmidrule(r){1-3}
Environment State Classification & Given an image, what are the lighting and climate conditions of it?& Visdrone-DET\cite{zhu2021detection}, Visdrone-VID\cite{zhu2021detection}, ERA\cite{mou2020era}, AIDER\cite{kyrkou2019deep}, MDOT\cite{zhu2020multi}\\
\cmidrule(r){1-3}
Urban OCR & Given an image, recognize the text within the selected region. & Visdrone-DET\cite{zhu2021detection}\\
\cmidrule(r){1-3}
Class-agnostic Counting & Count the number of the specified type of object in a given image.& Animaldrone\cite{zhu2021graph}, Cattle-det\cite{shao2020cattle}, MTC-plant\cite{lu2017tasselnet}\\
\cmidrule(r){1-3}
Referring Expression Counting & Count the number of objects that match the referring expression in a given image. & Refdrone\cite{sun2025refdrone}, Rec8k\cite{dai2024referring}, Visdrone-DET\cite{zhu2021detection}\\

\midrule % 粗线：Perception / Cognition 分组线
% \multicolumn{3}{@{}l}{\textbf{Cognition Tasks}} \\ % L1/L2 分组标题
% \midrule

Target Backtracking & Given multiple key frames in a video, what are the [spatial position] or [behavior] of the target object or person in the [past spatiotemporal sequence]?& Visdrone-VID\cite{zhu2021detection}, Visdrone-SOT\cite{zhu2021detection}\\
\cmidrule(r){1-3}
Cross-Object Reasoning & Given an image and multiple target objects or people, what are the [behavioral relationship] or [spatial relationship] between them?& Visdrone-DET\cite{zhu2021detection}\\
\cmidrule(r){1-3}
Intent Analysis and Prediction & Given multiple frames in a video and a target object or person, based on its [spatial position and behavior] [from the past to the present], what will its [future] [spatial position and behavior] be? & Mavrec\cite{dutta2024multiview}, MDOT\cite{zhu2020multi}\\
\cmidrule(r){1-3}
Scene Attribute Understanding & Given an image or a selected region within the image, does it conform to a certain [description]? / what is its [function]?& Visdrone-DET\cite{zhu2021detection}\\
\cmidrule(r){1-3}
Scene Damage Assessment & Given an image, how severe is the [disaster level] of the scene it shows? / Given multiple images, sort them by the severity of their disaster levels.& RescueNet\cite{rahnemoonfar2023rescuenet}\\
\cmidrule(r){1-3}
Scene Analysis and Prediction & Given multiple frames in a video, based on the changes the scene has undergone [from the past to the present]. What its [future change trend] be? & UAVid\cite{lyu2020uavid}, AU-AIR\cite{bozcan2020air}\\
\cmidrule(r){1-3}
Event Tracing & Given multiple frames in a video, analyze the [causes of the event] happened in the [past spatiotemporal sequence]?& ERA\cite{mou2020era}\\
\cmidrule(r){1-3}
Event Understanding & Given multiple frames in a video, understand the [event] that is [currently] happening.& ERA\cite{mou2020era}\\
\cmidrule(r){1-3}
Event Prediction & Given multiple frames in a video, based on the [event] happened [from the past to the present], predict the [future evolution] of the event.& ERA\cite{mou2020era}\\
\cmidrule(r){1-3}
Temporal Ordering & Given multiple [shuffled frames] of a video, based on the [stages of events] in different frames, what is the [correct chronological order] of the frames& ERA\cite{mou2020era}\\

\midrule % 粗线：Cognition / Planning 分组线
% \multicolumn{3}{@{}l}{\textbf{Planning Tasks}} \\ % L1/L2 分组标题
% \midrule

Swarm Collaborative Planning & Given multiple images from [multiple drone perspectives], select the perspective with the most comprehensive information as the main perspective and mark [multiple candidate regions in the main perspective]. Under a specific requirement, which region should [be prioritized for allocation]? / If the drone corresponding to the perspective of a certain image is [damaged], which region needs to have [a new drone added]?& MDOT\cite{zhu2020multi}\\
\cmidrule(r){1-3}
Ground-Target Planning & Given an image of a [disaster-affected scene] and [multiple marked rescue points], what is the [most suitable rescue route] and plan considering both rescue [priority] and rescue [time]?& AIDER\cite{kyrkou2019deep}\\
\cmidrule(r){1-3}
Air-Ground Collaborative Planning & Given multiple frames in a video, assuming [a task] needs to be performed, what actions should the [ground targets] and [UAVs] take respectively based on the current state?& ERA\cite{mou2020era}\\

\bottomrule
\end{tabular}

\end{table*}
% UAVBench 82%的任务来自于人工标注，剩下18%的任务是从人工标注的公开数据集中基于规则进行转换。对于人工标注的任务，我们首先会提供给标注员一份任务模板，标注员理解这个原子任务后在相应的数据集中挑选可标注的数据。每个任务的原型以及数据来源见表3.
\OURS consists of 82\% manually annotated tasks, while the remaining 18\% are automatically converted from publicly available dataset.

For manually annotated tasks, annotators are divided into three groups according to the L1 task categories. Each group is provided with a customized task template. After understanding the corresponding atomic task, annotators select suitable data instances from the collected datasets and start the annotation process.
The prototypes of each task and their associated data sources are summarized in Table~\ref{tab:template_source}.

% % \subsection{Quality Control Cases}
% % annotation accuracy : 放答疑文档+十字路口描述
% % task difficulty: 放event类总结的细节，标planning总结的细节
% \input{UAVBench/tables/quality_control}

The tasks derived from publicly datasets are Class-agnostic Counting, Referring Expression Counting, and Scene Damage Assessment.
For counting tasks, we convert raw count annotations into multiple-choice questions by generating distractors with controlled deviations determined by the difficulty setting (e.g., if the ground truth $c\in[20,50]$, distractors are generated as $c \pm 0.15c$, $c \pm 0.3c$. ). We then randomize the placement of the correct answer to avoid positional bias, ensuring that all options remain plausible while preserving fine-grained difficulty control.
% 对于Scene Damage Assessment，我们将原标注中对灾害程度描述的语义级信息加权计算成整张图的灾害数值。按照灾害数值分成不同的区间归类为"No Damage","Minor Damage", "Major Damage"和"Total Destruction"。这个任务分成对单张图的灾情程度判定和比较三张图的灾情严重程度。
For Scene Damage Assessment, we first established a quantitative metric by aggregating and weighting semantic damage descriptions from original annotations to derive a numerical damage score for each image. This score is mapped to four distinct severity levels ('No Damage', 'Minor Damage', 'Major Damage', and 'Total Destruction'). The task is divided into two sub-challenges: single-image assessment, where the model predicts the damage level of an individual image, and comparative ranking, where the model ranks the severity of damage across a set of three images.

% \subsubsection{Annotation workflows}
% FIG: construction pipeline
% The overall data construction pipeline of UAVBench is shown in Figure~\ref{fig:pipeline}. Here we provide the detail annotation workflows of each task. 
% \begin{itemize}
% \item \textbf{Scene Classification.}
% \end{itemize}

% \subsection{More statistics of UAVBench}
% FIG: word count
% \begin{figure}[t!]
% \centering
% \includegraphics[width=\linewidth]{UAVBench/figures/word_cloud.pdf}
% \caption{Word cloud of UAVBench.}
% \label{fig:pipeline}
% \end{figure}
% % FIG: scene diversity: 按照场景类别给case图
% \begin{figure}[t!]
% \centering
% \includegraphics[width=\linewidth]{UAVBench/figures/examples of dataset.pdf}
% \caption{Dataset examples from UAVBench.}
% \label{fig:pipeline}
% \end{figure}

\section{Evaluation Details.}
% \subsection{Experimental Setup}
We evaluate all models using VLMEvalKit\cite{duan2024vlmevalkit}. Our benchmark includes three input modalities: single image, key frames, and videos. For video-based tasks, the original videos are sampled at 3.0 fps. The evaluation prompt is provided below:

\begin{tcolorbox}[colback=black!5!white,colframe=black!75!black,title=Evaluation Prompt]

You are an expert in the field of drones. Please answer the following questions based on your professional knowledge.\\

\textit{For images or key frames input:} \\  
You have been provided with several images and a multiple-choice question related to the image. \\

\textit{For video input:} \\
You have been provided with \{len(frames)\} separate frames uniformly sampled from a video and a multiple-choice question related to the video. 
The frames are provided in chronological order of the video.\\

Your task is to carefully analyze the input data to answer the question, choosing from the options provided. Respond with only the letter of the correct option. \\

Question: \{Question\}

Options: \{Options\} \\

Please select the correct answer from the options above.

\end{tcolorbox}
We further use `exact\_matching' policy to extract response from the generated outputs. 
For efficiency in evaluating baseline models, we utilized vLLM\cite{vllm} to accelerate inference for models based on the Qwen architecture, while other models were run using the standard Hugging Face transformers library\cite{huggingface}. 
To ensure reproducibility, we strictly set the generation configuration with $temperature=0.0$ and $top\_p=1.0$. Furthermore, where supported by the model, we optionally employed $num\_beams=3$ for generation.

% \subsection{Introduction of Baseline}
% % 对于Qwenvl 系列，采用vllm评测
% % 对于Internvl 系列，
% We introduce the MLLMs for zero-shot evaluation as follows:

% Gemini-

% Part 2
\section{Extended Results and Analysis.}
Here we provide more evaluation results and analysis on \OURS, including L2 category results, results of models with Chain-of-Thought(CoT), and more analysis on spacial prediction bias.

% \input{UAVBench/tables/blind_evaluation}
% % 还有张图说明差距
% % 结果太差就不放了
\begingroup
\definecolor{perceptionColor}{HTML}{96A9BF} 
\definecolor{cognitionColor}{HTML}{EBD8B7}
\definecolor{planningColor}{HTML}{B5A69C}   
\definecolor{oai-gray-600}{RGB}{232, 159, 44}
\definecolor{oai-gray-300}{RGB}{254, 210, 103}
\definecolor{oai-red-1}{RGB}{250, 40, 55}
\definecolor{oai-red-2}{RGB}{255, 110, 120}
\definecolor{oai-red-3}{RGB}{247, 198, 198}

\begin{table}[ht!]
    \caption{Averaged zero-shot evaluation results on \OURS on L2 Category.}
    \vspace{-0.3em}
    \centering
    \fontsize{6.5pt}{7.5pt}\selectfont
    \setlength\tabcolsep{3pt}
    \renewcommand{\arraystretch}{1.5}
    \begin{tabular}{r|cccccccc}
         & \rotatebox{75}{Classification} & \rotatebox{75}{OCR} & \rotatebox{75}{Counting} & \rotatebox{75}{Object} & \rotatebox{75}{Scene} & \rotatebox{75}{Event} & \rotatebox{75}{UAV-to-UAV} & \rotatebox{75}{UAV-to-Ground} \\
         Methods & 
        \multicolumn{3}{c}{\cellcolor{perceptionColor!50}Perception} & 
        \multicolumn{3}{c}{\cellcolor{cognitionColor!50}Cognition} & 
        \multicolumn{2}{c}{\cellcolor{planningColor!50}Planning} \\
        % \hline
        % \rowcolor{gray!10}
        % \multicolumn{1}{l|}{\textcolor{black}{\textit{Baseline}}} & & & & & & & & \\
        % Random & 25.00 & 25.00 & 20.00 & 26.54 & 27.52 & 24.78 & 28.07 & 23.96 \\
        % Human{$^\ddag$} & 82.06 & 76.67 & 35.92 & 86.78 & 88.36 & 90.29 & 85.71 & 80.31 \\
        \hline
        \rowcolor{gray!10}
        \multicolumn{1}{l|}{\textcolor{black}{\textit{API-based}}} & & & & & & & & \\
        Gemini 2.5 Pro & 62.19 & 82.19 & 24.40 & 50.05 & 61.89 & 66.62 & 25.68 & 46.38 \\
        Gemini 2.5 Flash & 60.00 & 75.94 & 28.71 & 37.62 & 57.27 & 56.44 & 15.54 & 30.07 \\
        GPT-4o & 54.59 & 62.19 & 20.49 & 30.00 & 51.74 & 57.18 & 28.38 & 42.14 \\
        \hline
        \rowcolor{gray!10}
        \multicolumn{1}{l|}{\textcolor{black}{\textit{Open-source}}} & & & & & & & & \\
        Qwen3-VL-8B & 56.05 & 79.69 & 28.68 & 55.25 & 57.57 & 53.11 & 27.12 & 42.76 \\
        Qwen3-VL-32B & 61.88 & 76.56 & 36.01 & 48.90 & 63.87 & 61.96 & 37.63 & 47.19 \\
        Qwen3-VL-235B-A22B & 61.99 & 73.44 & 36.64 & 48.97 & 58.09 & 62.84 & 33.45 & 48.85 \\
        Qwen2.5-VL-7B & 55.82 & 70.31 & 28.86 & 49.34 & 56.22 & 51.97 & 27.12 & 45.42 \\
        Qwen2.5-VL-32B & 55.15 & 76.56 & 36.01 & 45.69 & 64.69 & 56.59 & 23.73 & 48.62 \\
        Qwen2.5-VL-72B & 58.94 & 75.62 & 29.87 & 48.82 & 59.85 & 66.21 & 35.25 & 49.16 \\
        InternVL3.5-8B & 52.61 & 68.12 & 27.04 & 40.12 & 57.02 & 51.34 & 40.68 & 38.34 \\
        InternVL3.5-38B & 41.38 & 66.56 & 33.69 & 33.43 & 46.61 & 56.31 & 32.54 & 32.23 \\
        InternVL3-14B & 52.33 & 73.75 & 24.36 & 43.12 & 61.53 & 51.58 & 29.49 & 35.44 \\
        InternVL3-78B & 63.96 & 58.44 & 30.73 & 45.59 & 65.43 & 63.38 & 38.64 & 40.40 \\
        LLaVA-OneVision-7B & 50.21 & 65.94 & 24.72 & 37.26 & 38.40 & 51.09 & 27.46 & 44.00 \\
        MiniCPM-V-4.5-8B & 57.17 & 72.50 & 32.96 & 43.66 & 43.14 & 52.56 & 39.32 & 43.03 \\
        MiMo-VL-7B-RL & 53.49 & 70.62 & 24.20 & 41.28 & 55.89 & 39.19 & 23.73 & 45.38 \\
        \hline
    \end{tabular}
    \label{tab:recategorized_results_v2_simple_no_rank}
\end{table}
\endgroup 
\subsection{Results on L2 Category}
% 模型在L2 category上的表现更加反映当前MLLMs对无人机场景缺少的关键能力。
% 在Perception维度考察的三种能力上，所有模型的Counitng能力显著弱于Classification和OCR能力。
%在Cognition维度考察的不同对象的认知推理上，模型在Object的认知推理上明显弱于scene level和event level。可能是源于object的target scale太小，感知难度比scene和event更难，所以认知难度也更难。这里可以在正文实验分析4.3节图4中得到论证，认知任务Target Backtracking的性能随着target sizes增大而性能上升。
% 在Planning维度考察的中，MLLMs在UAV-to-UAV的表现比UAV-to-Ground差，在数据集输入上UAV-to-UAV是空中多视角图像，UAV-to-Ground输入都是单一视角图像，可能源自MLLMs无法很好的处理多视角问题。

Table~\ref{tab:recategorized_results_v2_simple_no_rank} presents the zero-shot evaluation results of 16 general MLLMs across the L2 categories, which directly reveals specific capability deficiencies required for UAV scenarios. We can summarize limitations of MLLMs across three core dimensions:

\begin{itemize}
    \item[$\bullet$] \textbf{ Weakness in fine-grained quantitative perception.}
    In the Perception dimension, performance on Counting ($\sim 20-36\%$) is significantly lower than Classification and OCR ($\sim 50-80\%$). This disparity underscores a severe bottleneck where MLLMs struggle with fine-grained enumeration and density estimation in aerial imagery.
    
    \item[$\bullet$] \textbf{Object-level reasoning is more challenging.}
    The Cognition dimension reveals that object-level reasoning is notably weaker than both scene-level and event-level reasoning. This decline is possibly correlated with the small target scale and limited context. 
    Fig.~\ref{fig:object_scale} also shows that Target Backtracking (an object-level reasoning task) improves with increasing target size.
    
    \item[$\bullet$] \textbf{Handling multi-view images increases the difficulty of planning.} 
    MLLMs perform substantially worse on UAV-to-UAV collaborative tasks (multi-view input) than on UAV-to-Ground tasks (single-view input) in the Planning dimension.
    This strongly indicates MLLMs’ fundamental limitation in processing and integrating information from disparate multi-view inputs, which is critical for complex swarm planning.
\end{itemize}

\subsection{Evaluation with CoT}
\begingroup
% \definecolor{perceptionColor}{HTML}{87CEEB} 
\definecolor{perceptionColor}{HTML}{96A9BF} 
% \definecolor{cognitionColor}{HTML}{DDA0DD}   
\definecolor{cognitionColor}{HTML}{EBD8B7}
% \definecolor{planningColor}{HTML}{FFA07A}   
\definecolor{planningColor}{HTML}{B5A69C}   
% \definecolor{oai-gray-600}{RGB}{130, 130, 130}
% \definecolor{oai-gray-600}{RGB}{108, 100, 158}
\definecolor{oai-gray-600}{RGB}{232, 159, 44}
% \definecolor{oai-gray-300}{RGB}{220, 220, 220}  
% \definecolor{oai-gray-300}{RGB}{173, 168, 208} 
\definecolor{oai-gray-300}{RGB}{254, 210, 103}
\definecolor{oai-red-1}{RGB}{250, 40, 55}
\definecolor{oai-red-2}{RGB}{255, 110, 120}
\definecolor{oai-red-3}{RGB}{247, 198, 198}

% 导言区需提前加载：\usepackage{graphicx}、\usepackage{xcolor}
\begin{table*}[ht!]  % 自动占满双栏宽度
    \captionsetup{type=table}
    \caption{\textbf{Evaluation with CoT on L3 sub-tasks.} ``Qwen3-VL-8B-Thinking" and ``MiMo-VL-7B-RL-Thinking" denote the original models augmented with CoT. $\Delta$ represents the performance difference (CoT-augmented minus original) for each sub-task. }
    \vspace{-0.3em}
    \centering
    \fontsize{6pt}{7pt}\selectfont  % 增大字体提升可读性
    \setlength\tabcolsep{3pt}  % 拉宽列间距（原2pt→4pt，可按需微调）
    \renewcommand{\arraystretch}{1.5}  % 增大行高，避免文字拥挤
    \begin{tabular}{r|c|cccccccccccccccccccc}
       & &
        \rotatebox{75}{Scene. Class.} &
        \rotatebox{75}{Orient. Class.} &
        \rotatebox{75}{Env. State} & 
        \rotatebox{75}{Urban OCR} &
        \rotatebox{75}{CA. Count} &
        \rotatebox{75}{RE. Count} &
        \rotatebox{75}{Target Back.} &
        \rotatebox{75}{Cross-Obj R.} &
        \rotatebox{75}{Intent Anal.} &
        \rotatebox{75}{Scene Attri.} &
        \rotatebox{75}{Scene Damage} &
        \rotatebox{75}{Scene Pred.} &
        \rotatebox{75}{Event Trace} &
        \rotatebox{75}{Event Under.} &
        \rotatebox{75}{Event Pred.} &
        \rotatebox{75}{Temporal Order} &
        \rotatebox{75}{Swarm Plan} &
        \rotatebox{75}{Ground Plan} &
        \rotatebox{75}{Air-Ground Plan} \\
        Methods & Avg. & 
        \multicolumn{6}{c}{\cellcolor{perceptionColor!50}Perception} & 
        \multicolumn{10}{c}{\cellcolor{cognitionColor!50}Cognition} & 
        \multicolumn{3}{c}{\cellcolor{planningColor!50}Planning} \\
        \hline    
        Qwen3-VL-8B & 50.98 & 69.87 & 26.52 & 71.76 & 79.69 & 38.79 & 18.57 & 50.00 & 57.56 & 58.18 & 85.56 & 34.00 & 53.14 & 58.44 & 62.46 & 61.54 & 30.00 & 33.11 & 45.61 & 39.90 \\
        Qwen3-VL-8B-Thinking & 53.54 & 69.46 & 45.20 & 71.47 & 72.50 & 47.88 & 24.56 & 40.17 & 43.02 & 52.68 & 84.06 & 41.33 & 66.67 & 63.79 & 65.78 & 65.59 & 39.47 & 36.21 & 46.31 & 41.06 \\
        $\Delta$ & \poscolor{+2.55} & \negcolor{-0.41} & \poscolor{+18.68} & \negcolor{-0.29} & \negcolor{-7.19} & \poscolor{+9.09} & \poscolor{+5.99} & \negcolor{-9.83} & \negcolor{-14.54} & \negcolor{-5.50} & \negcolor{-1.50} & \poscolor{+7.33} & \poscolor{+13.53} & \poscolor{+5.35} & \poscolor{+3.32} & \poscolor{+4.05} & \poscolor{+9.47} & \poscolor{+3.10} & \poscolor{+0.70} & \poscolor{+1.16} \\
        Mimo-VL-7B-RL & 44.67 & 67.36 & 24.22 & 68.88 & 70.63 & 16.97 & 31.43 & 43.22 & 32.56 & 48.06 & 84.48 & 29.14 & 54.11 & 40.74 & 39.20 & 36.84 & 40.00 & 30.08 & 48.44 & 42.31 \\
        Mimo-VL-7B-RL-Thinking & 50.22 & 67.36 & 35.05 & 69.74 & 70.31 & 25.15 & 32.27 & 31.62 & 49.42 & 47.05 & 86.28 & 35.82 & 53.47 & 64.73 & 63.33 & 54.29 & 36.98 & 41.02 & 45.58 & 44.66 \\
        $\Delta$ & \poscolor{+5.55} & \negcolor{-0.00} & \poscolor{+10.83} & \poscolor{+0.86} & \negcolor{-0.32} & \poscolor{+8.18} & \poscolor{+0.84} & \negcolor{-11.60} & \poscolor{+16.86} & \negcolor{-1.01} & \poscolor{+1.80} & \poscolor{+6.68} & \negcolor{-0.64} & \poscolor{+23.99} & \poscolor{+24.13} & \poscolor{+17.45} & \negcolor{-3.02} & \poscolor{+10.94} & \negcolor{-2.86} & \poscolor{+2.35} \\
        \hline
    \end{tabular}
    \label{tab:zero_shot_cot}
\end{table*}
\endgroup
% we further provide zero-shot evaluation on models with CoT. The performance gap between w and w/o CoT is shown in Tab.~\ref{tab:zero_shot_cot}. 首先，加了 CoT会在总体上提升模型性能，具体任务来看，

The performance gap between models with and without Chain-of-Thought (CoT) prompting is shown in Table~\ref{tab:zero_shot_cot}.
Overall, employing CoT significantly improves the average performance of both models (Qwen3-VL-8B:$\Delta +2.55$; MiMo-VL-7B-RL:$\Delta +5.55$). Across individual tasks, the gains are highly variable. 
CoT largely improves the performance in the perceptional tasks like Orientation Classification, Class-agnostic Counting, and most Event-Level tasks in cognition.  
Conversely, CoT does not perform well in Object-Level Reasoning tasks (e.g., Qwen3-VL-8B: $\Delta -14.54$), suggesting that the explicit intermediate steps may introduce errors when the initial perception or localization of the target is inherently difficult.

\subsection{Analysis on Spacial Prediction Bias}
% FIG: spacial prediction bias
\begin{figure*}[t!] % Use figure* for a double-column figure
    \centering
    % \begin{subfigure}{0.32\linewidth}
    %     \includegraphics[width=\linewidth]{UAVBench/figures/spatial_bias/all/heatmap_Qwen3-VL-8B.pdf}
    %     \caption*{(a) Qwen3-VL-8B}
    % \end{subfigure}
    \begin{subfigure}{0.32\linewidth}
        \includegraphics[width=\linewidth]{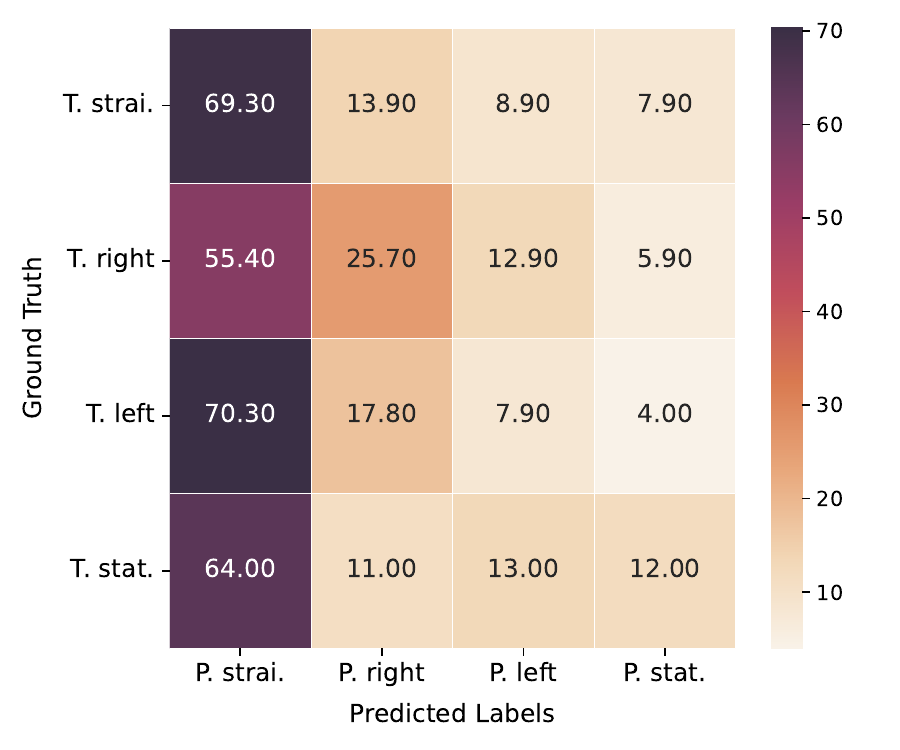}
        \caption*{(a) Gemini 2.5 Flash}
    \end{subfigure}
    \hfill
    \begin{subfigure}{0.32\linewidth}
        \includegraphics[width=\linewidth]{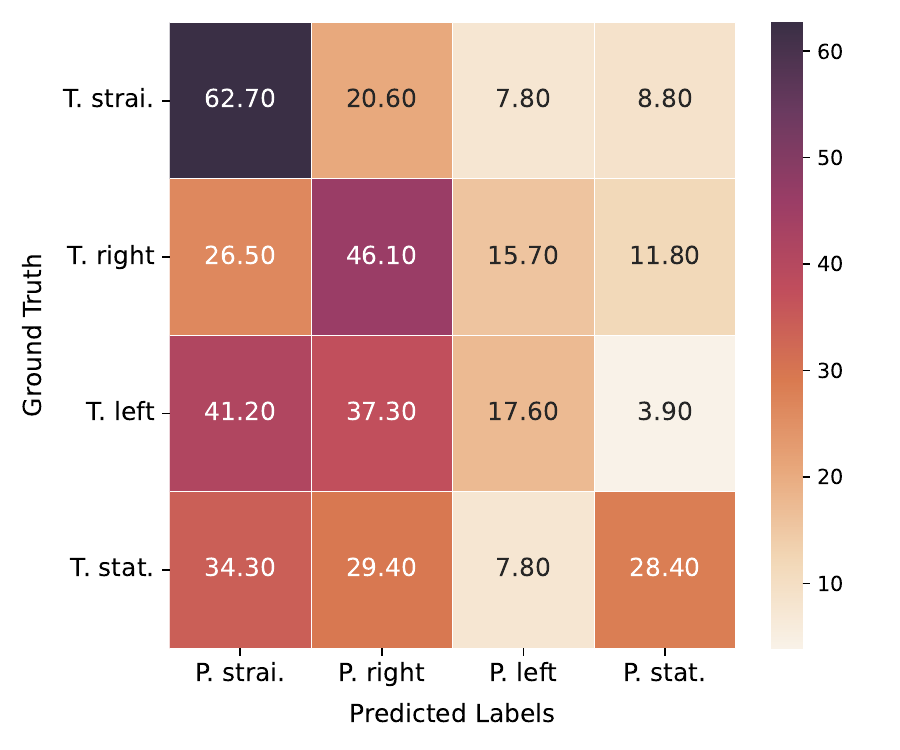}
        \caption*{(b) Qwen3-VL-32B}
    \end{subfigure}
    \hfill
    \begin{subfigure}{0.32\linewidth}
        \includegraphics[width=\linewidth]{figures/spatial_bias/heatmap_MiniCPM.png}
        \caption*{(c) Qwen3-VL-235B-A22B}
    \end{subfigure}

    \vspace{1em} % Add some vertical space between rows

    \begin{subfigure}{0.32\linewidth}
        \includegraphics[width=\linewidth]{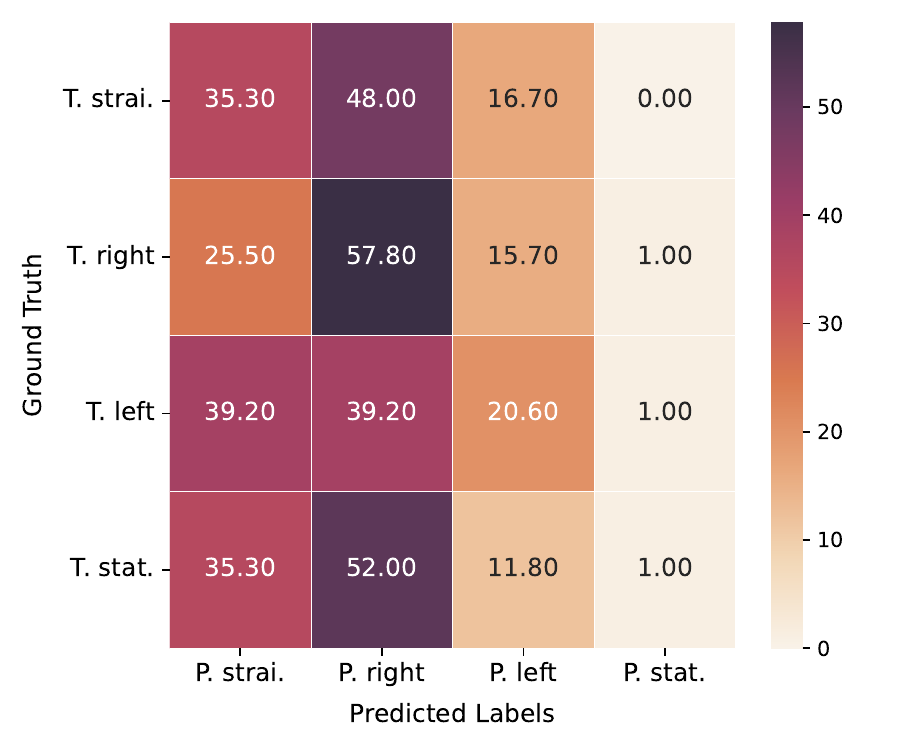}
        \caption*{(d) InternVL3.5-8B}
    \end{subfigure}
    \hfill
    \begin{subfigure}{0.32\linewidth}
        \includegraphics[width=\linewidth]{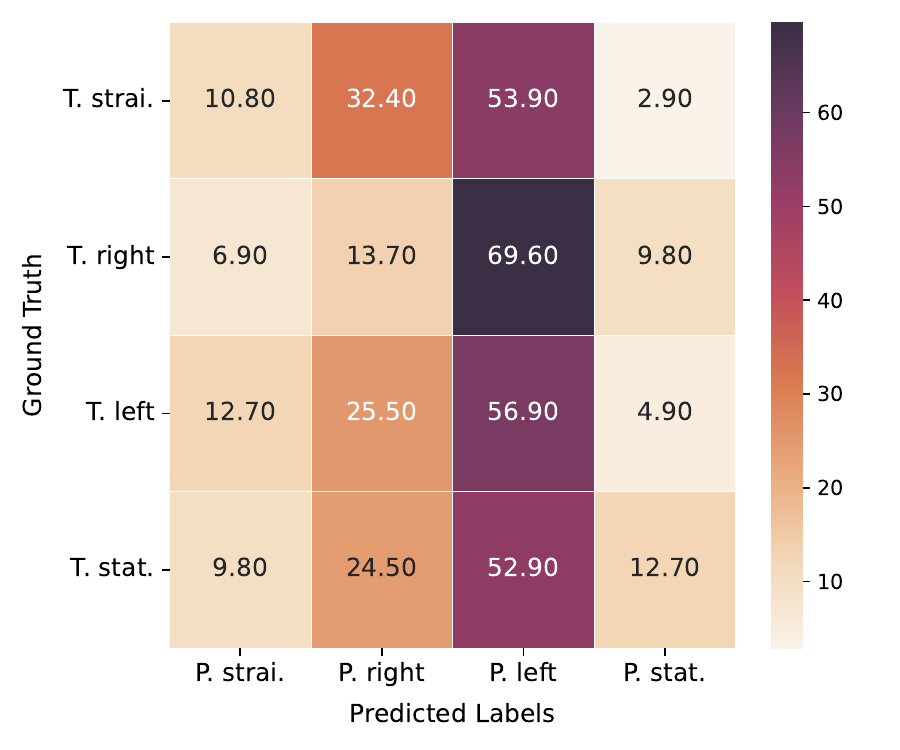}
        \caption*{(e) InternVL3-14B}
    \end{subfigure}
    \hfill
    \begin{subfigure}{0.32\linewidth}
        \includegraphics[width=\linewidth]{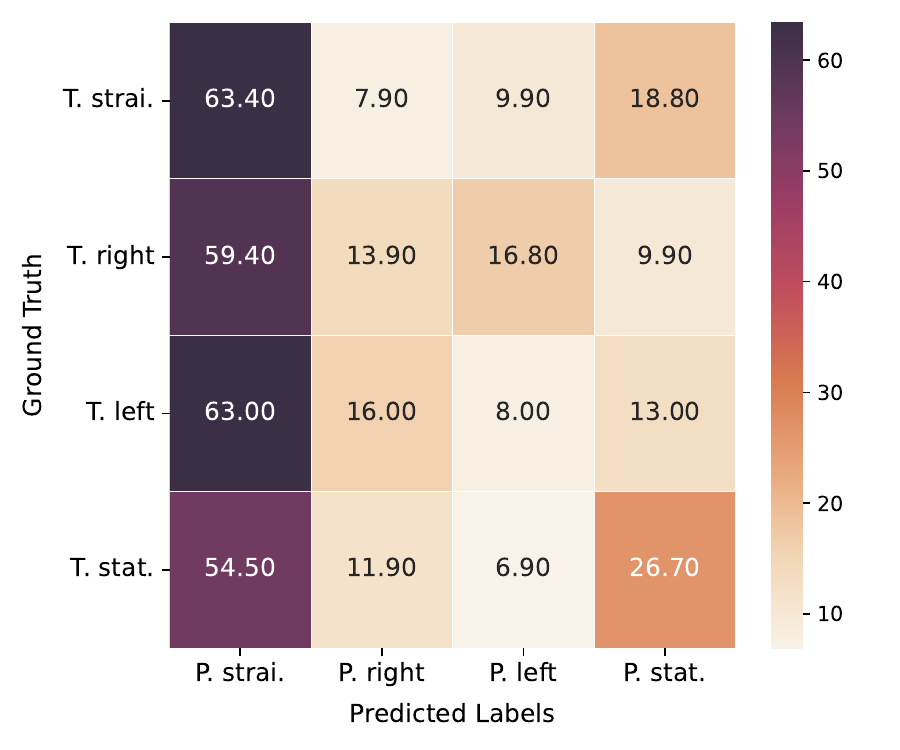}
        \caption*{(f) InternVL3.5-38B}
    \end{subfigure}
    
    \vspace{1em} % Add some vertical space between rows

    \begin{subfigure}{0.32\linewidth}
        \includegraphics[width=\linewidth]{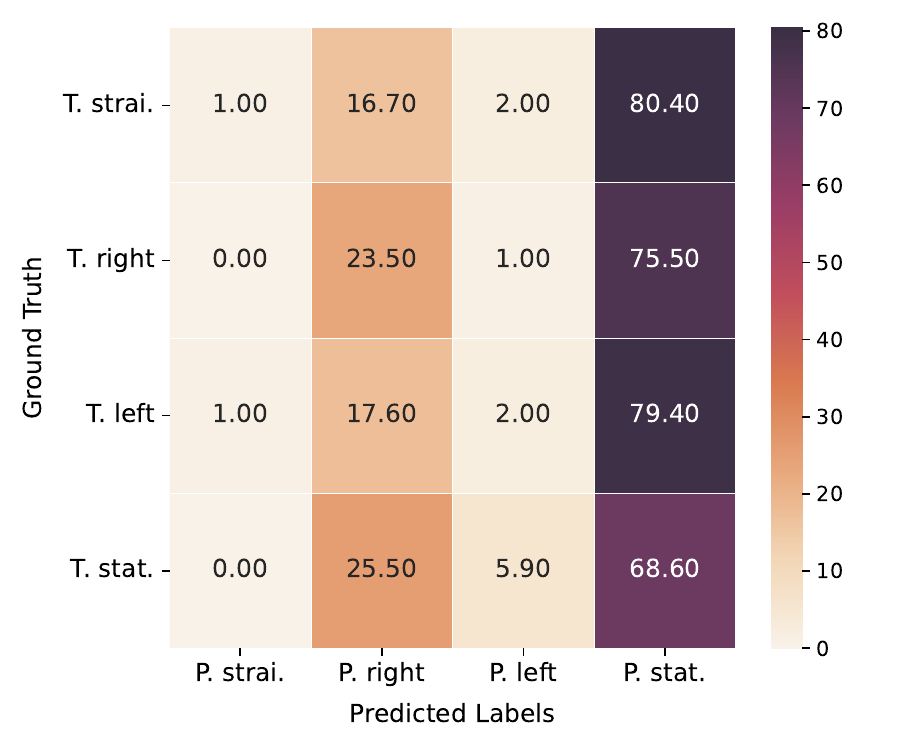}
        \caption*{(g) LLaVA-OneVision-7B}
    \end{subfigure}
    \hfill
    \begin{subfigure}{0.32\linewidth}
        \includegraphics[width=\linewidth]{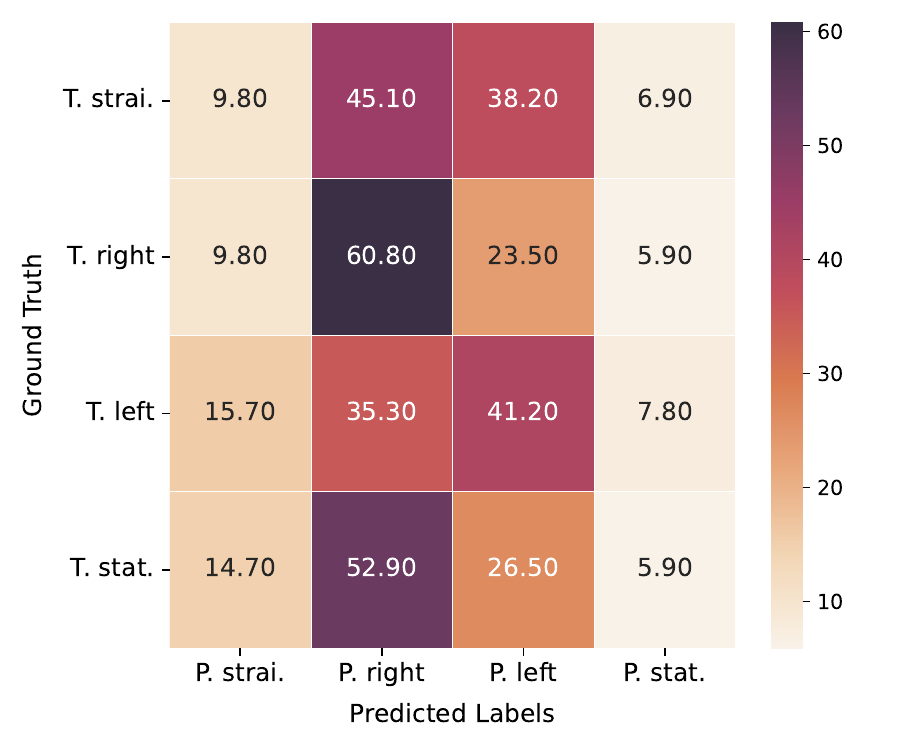}
        \caption*{(h) MiMo-VL-7B-RL}
    \end{subfigure}
    \hfill
    \begin{subfigure}{0.32\linewidth}
        \includegraphics[width=\linewidth]{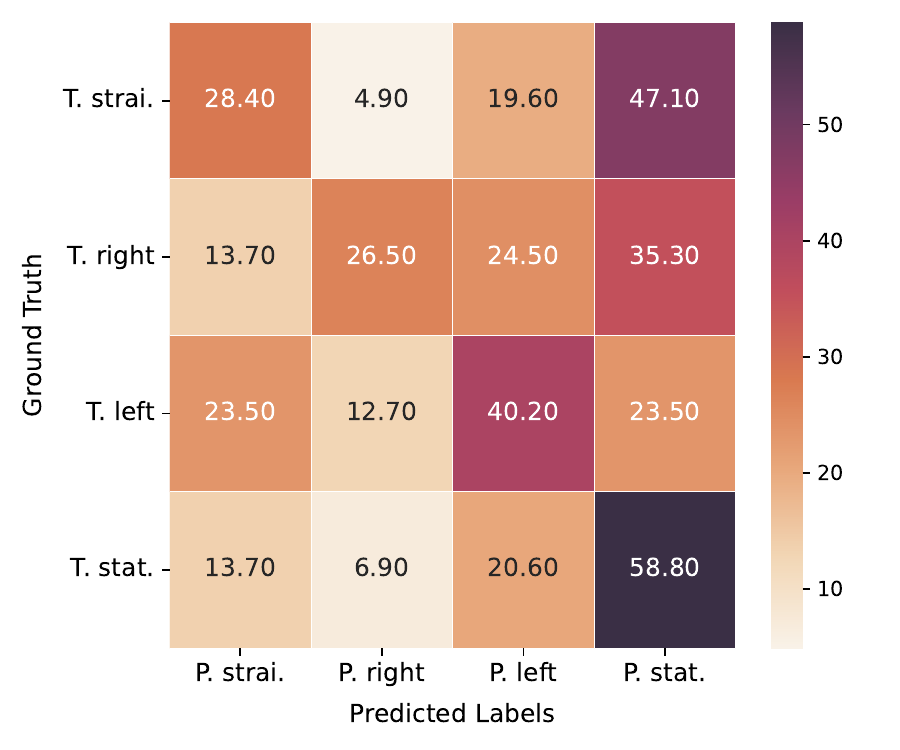}
        \caption*{(i) Gemini 2.5 Pro}
    \end{subfigure}
    \caption{ More confusion matrices of other baseline models on the Orientation Classification task.}
    \label{fig:spatial_bias_doublecol}
\end{figure*}

As shown in Fig.~\ref{fig:spatial_bias_doublecol}, we present the confusion matrices of other baseline models for the Orientation Classification task. 
Both API-based and open-source MLLMs exhibit spatial prediction biases. Specifically, models demonstrate a strong conservative bias, frequently misclassifying turning motions (`T. right', `T. left') as the `P. stat.' (predicted stationary) or `T. strai.' (predicted straight) categories. This indicates that MLLMs struggle to accurately resolve subtle directional cues in aerial imagery, leading to confusion between rotational and static/forward movement intentions.

% \subsection{More Cases on Error analysis}
% % FIG: error analysis 标在10里

% Part 3
\section{Visualizations and Challenging Cases}
\begin{figure*}[t]
  \centering
  \includegraphics[width=\textwidth]{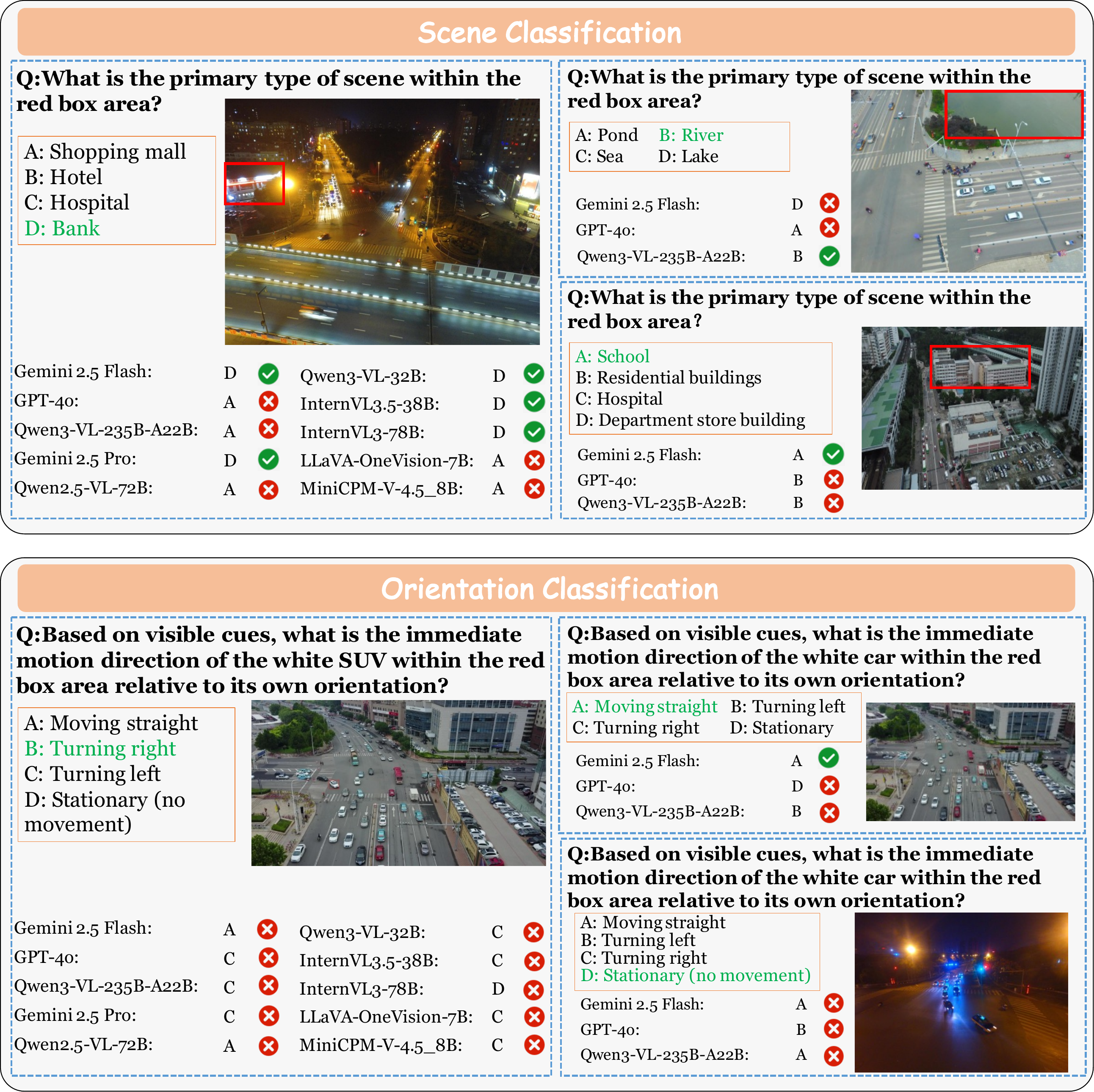} 
  \caption{Additional Examples of Sub-Tasks (Part 1).}
  \label{fig:t1t2}
\end{figure*}

\begin{figure*}[t]
  \centering
  \includegraphics[width=\textwidth]{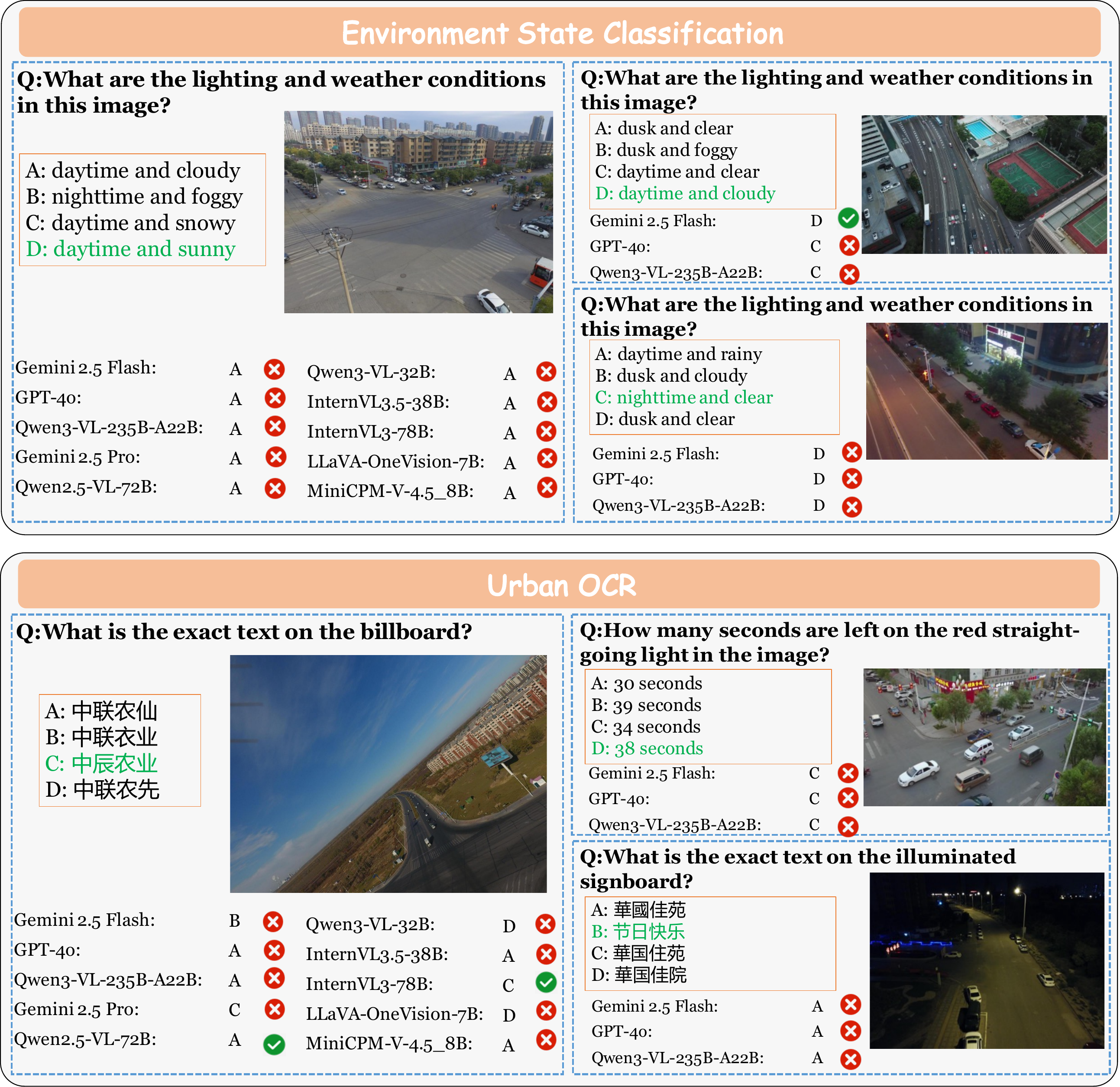} 
  \caption{Additional Examples of Sub-Tasks (Part 2).}
  \label{fig:t3t4}
\end{figure*}

\begin{figure*}[t]
  \centering
  \includegraphics[width=\textwidth]{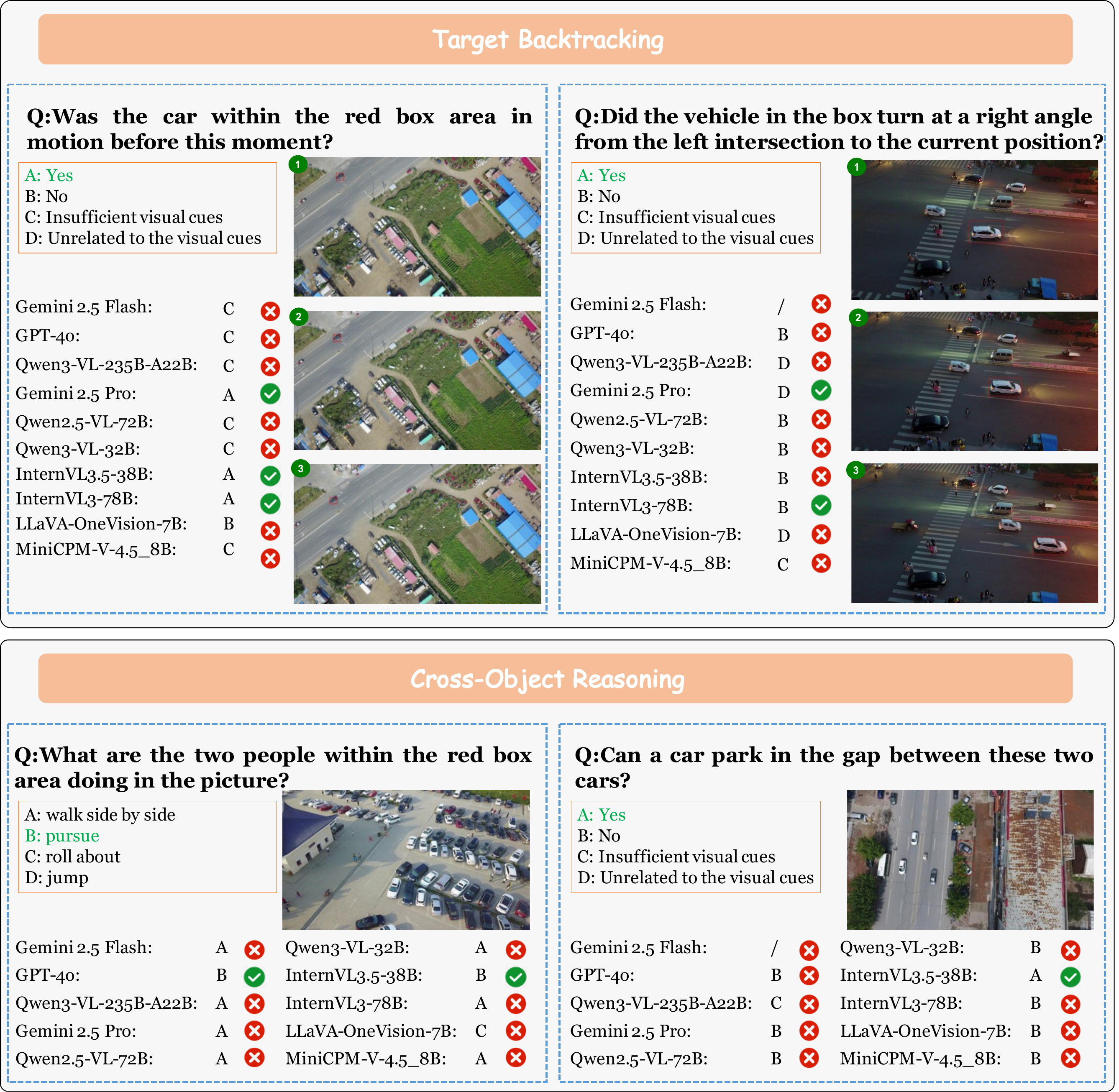} 
  \caption{Additional Examples of Sub-Tasks (Part 3).}
  \label{fig:t8t9}
\end{figure*}

\begin{figure*}[t]
  \centering
  \includegraphics[width=\textwidth]{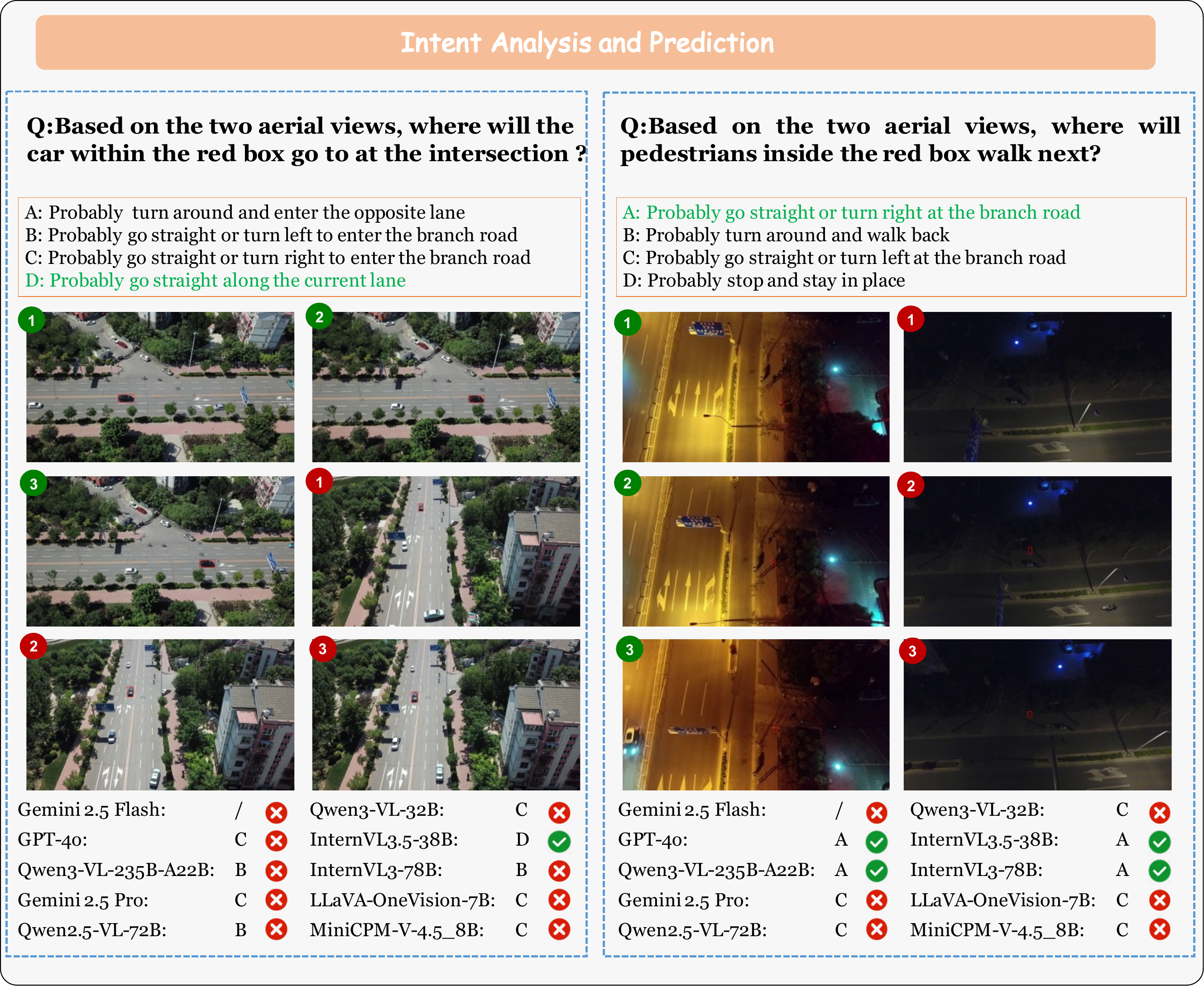} 
  \caption{Additional Examples of Sub-Tasks (Part 4).}
  \label{fig:t10}
\end{figure*}

\begin{figure*}[t]
  \centering
  \includegraphics[width=\textwidth]{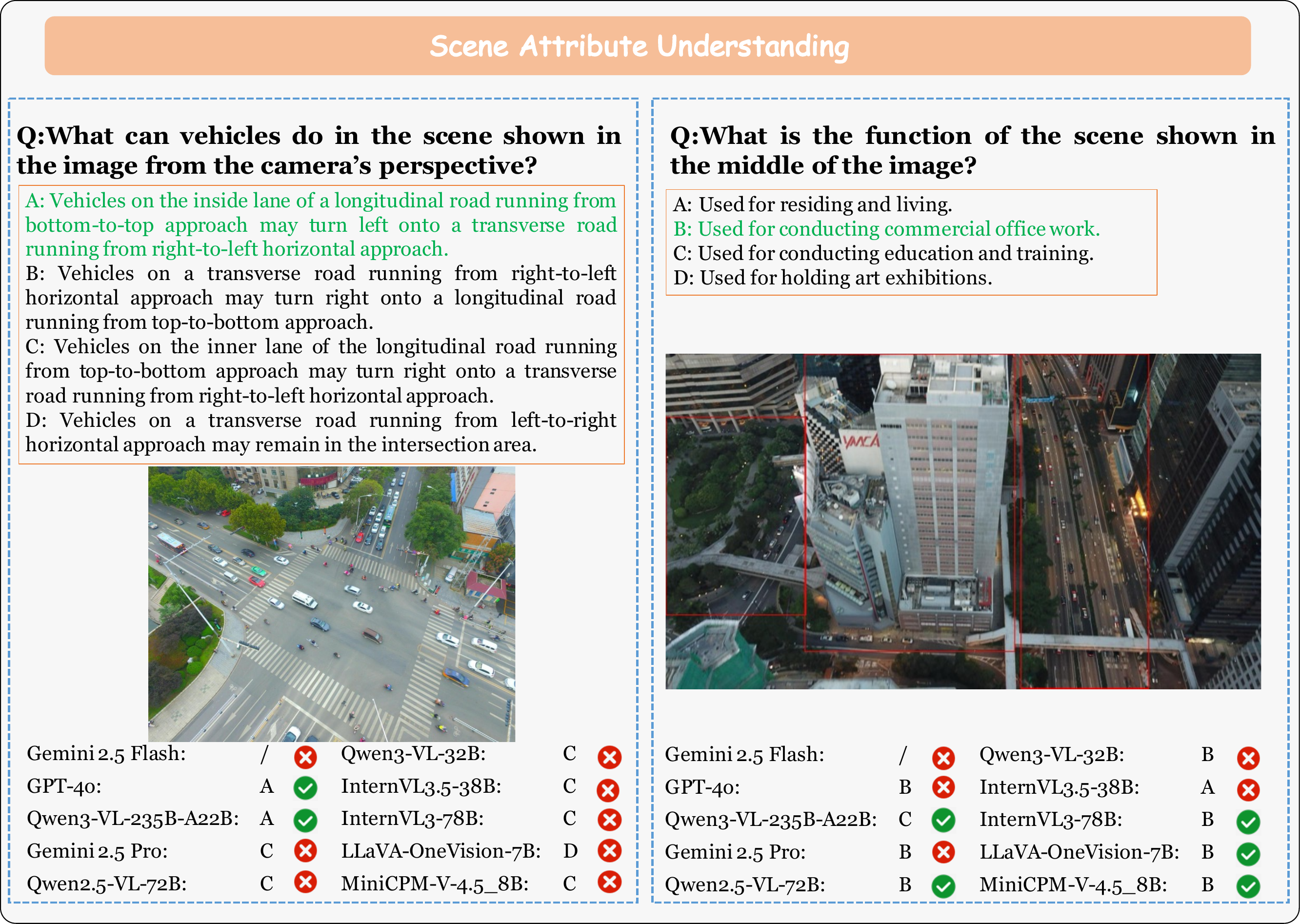} 
  \caption{Additional Examples of Sub-Tasks (Part 5).}
  \label{fig:t11}
\end{figure*}

\begin{figure*}[t]
  \centering
  \includegraphics[width=0.9\textwidth]{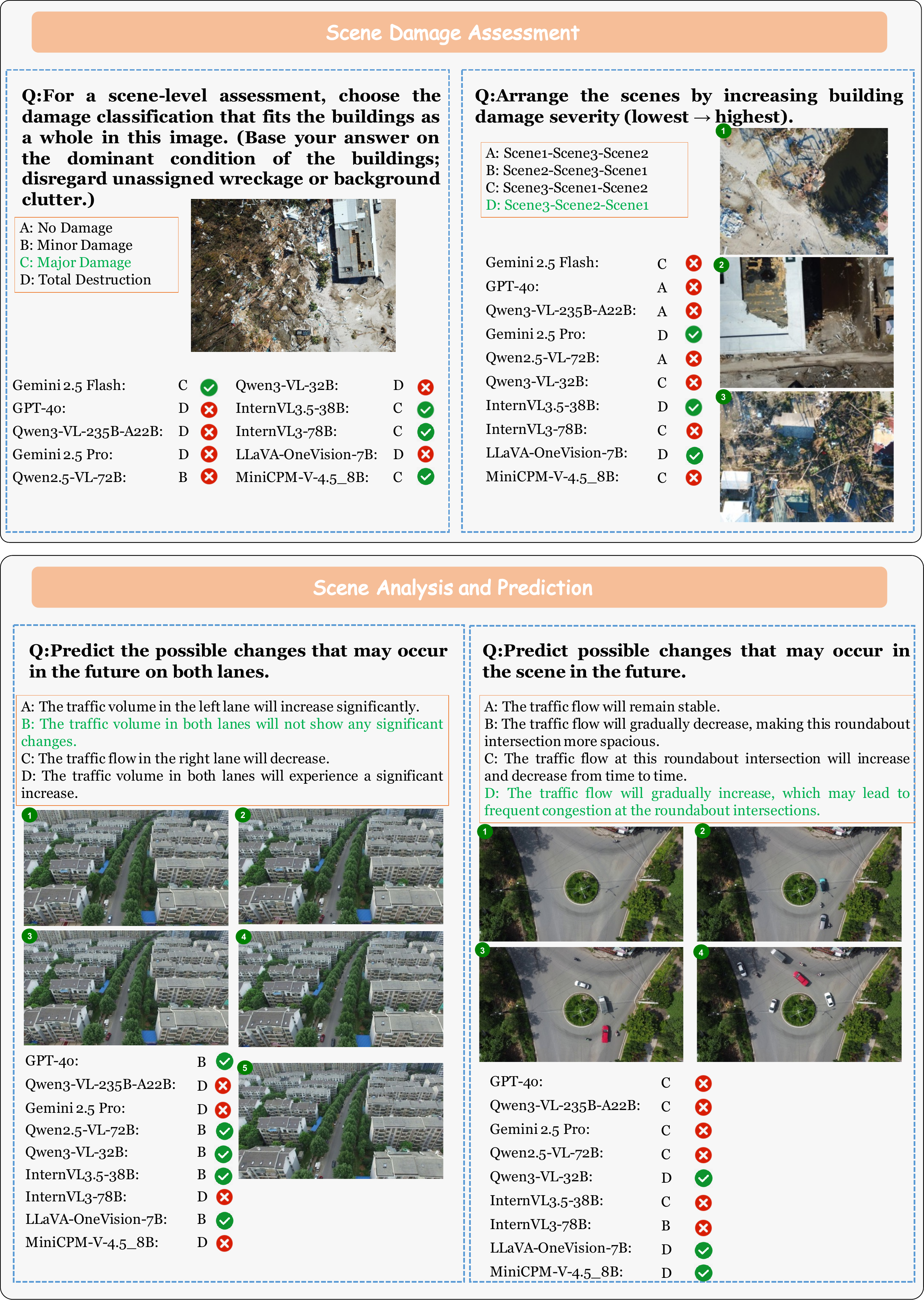} 
  \caption{Additional Examples of Sub-Tasks (Part 6).}
  \label{fig:t12t13}
\end{figure*}

\begin{figure*}[t]
  \centering
  \includegraphics[width=\textwidth]{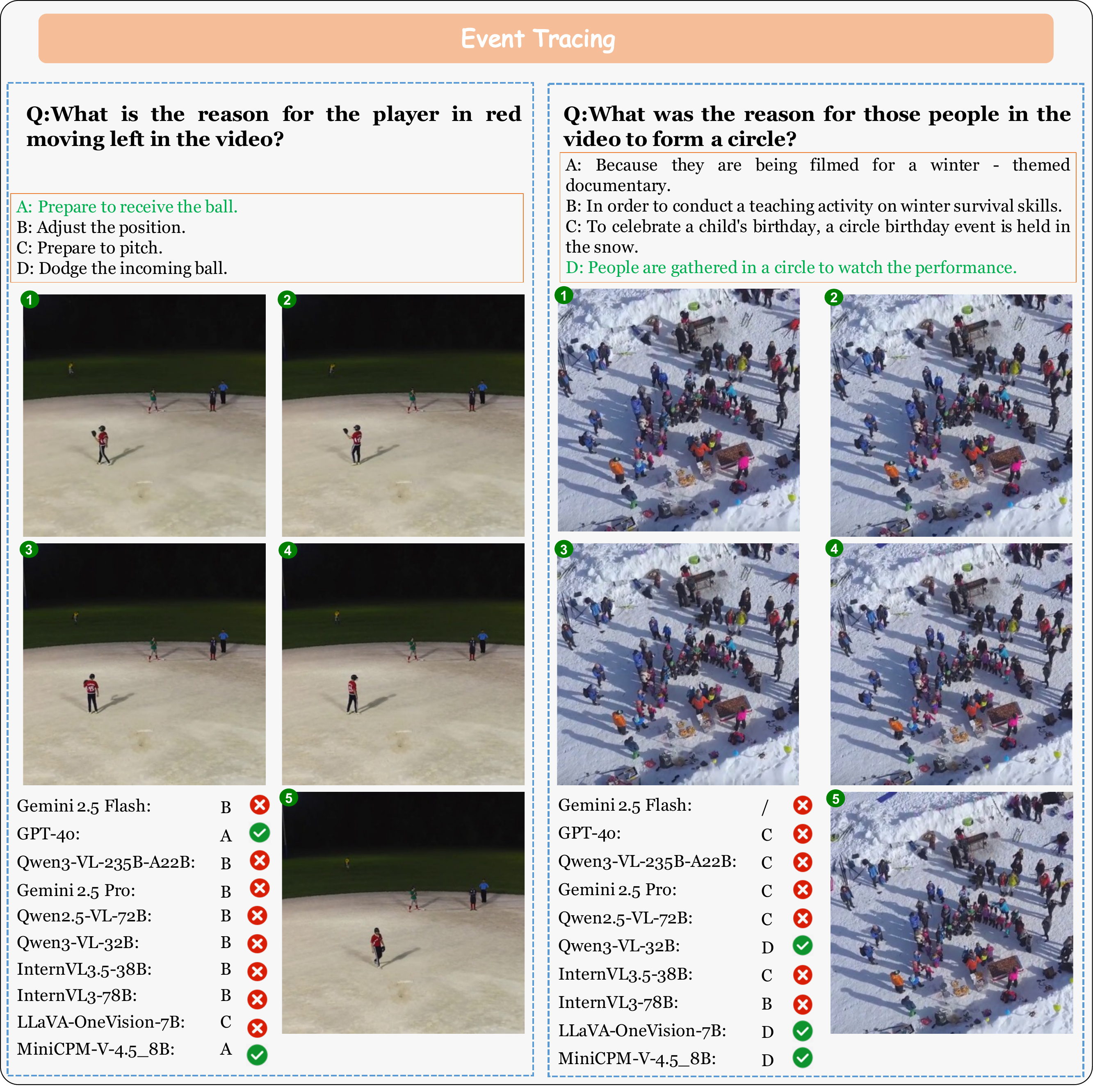} 
  \caption{Additional Examples of Sub-Tasks (Part 7).}
  \label{fig:t14}
\end{figure*}

\begin{figure*}[t]
  \centering
  \includegraphics[width=\textwidth]{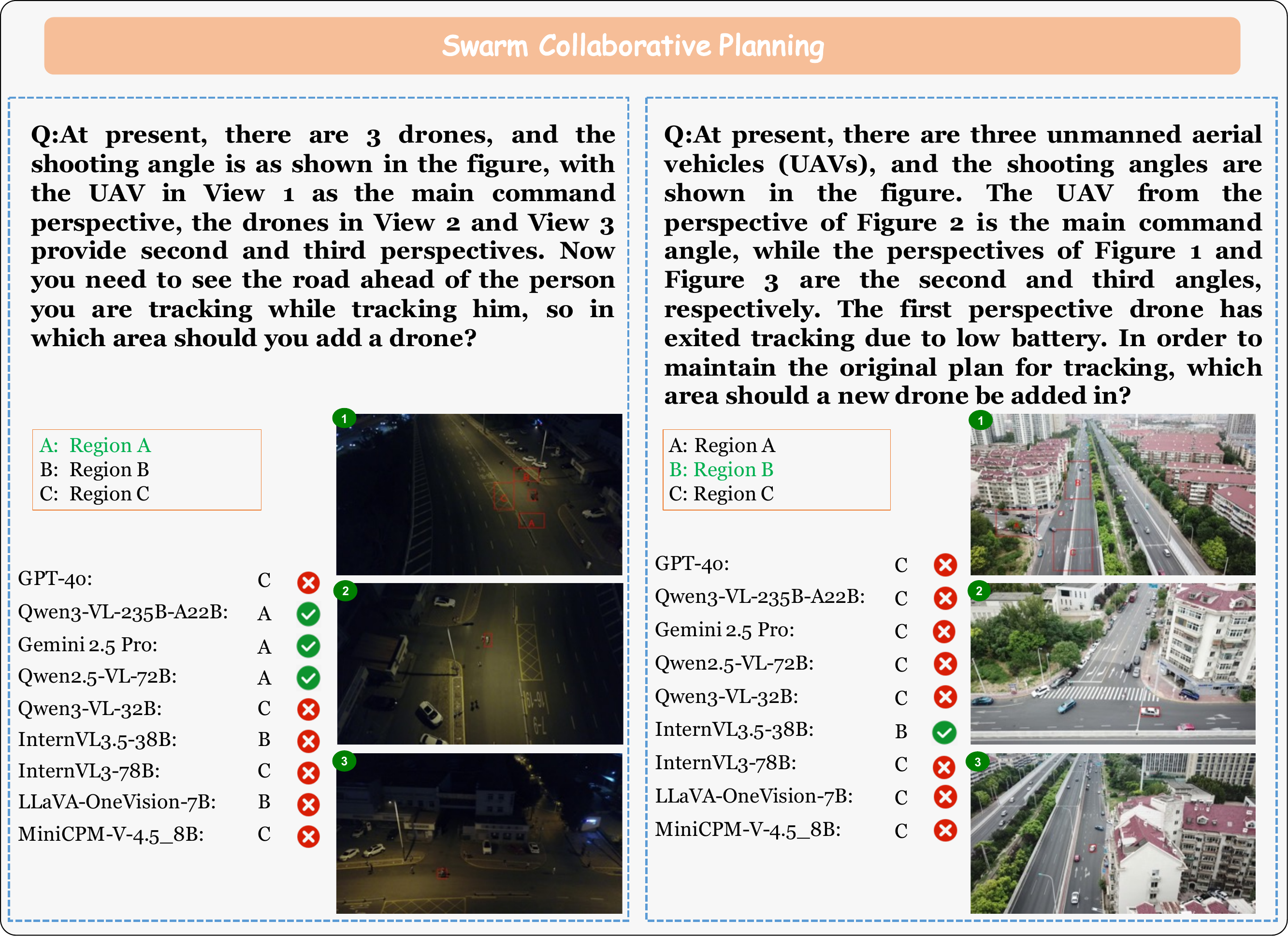} 
  \caption{Additional Examples of Sub-Tasks (Part 8).}
  \label{fig:t18}
\end{figure*}

\begin{figure*}[t]
  \centering
  \includegraphics[width=\textwidth]{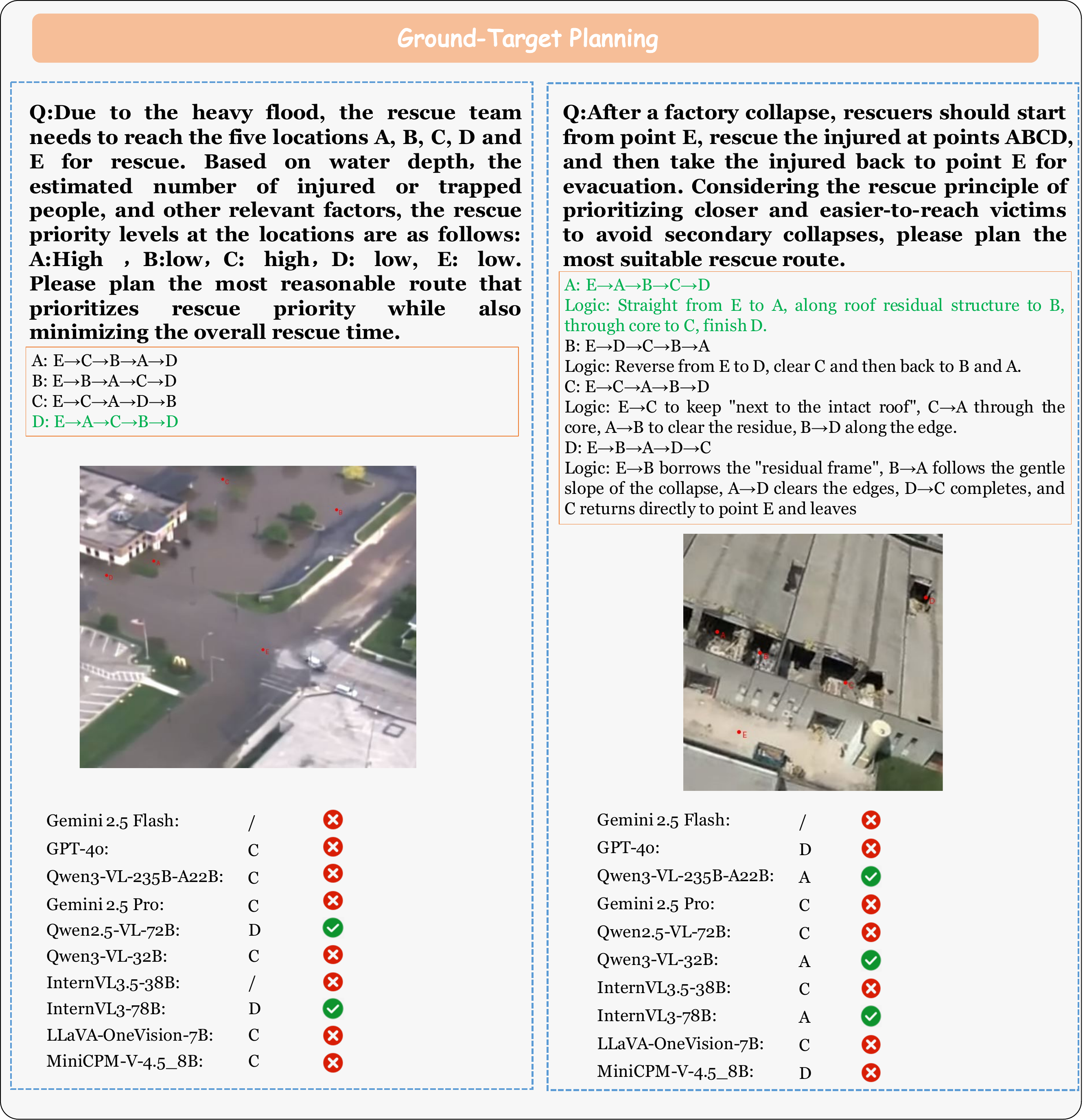} 
  \caption{Additional Examples of Sub-Tasks (Part 9).}
  \label{fig:t19}
\end{figure*}

In this section, we present additional examples from \OURS along with the responses of baseline models. 
Representative cases for the Perception dimension are shown in Fig.~\ref{fig:t1t2} and Fig.~\ref{fig:t3t4}, 
for the Cognition dimension in Fig.~\ref{fig:t8t9}, Fig.~\ref{fig:t10}, Fig.~\ref{fig:t11}, Fig.~\ref{fig:t12t13} and Fig.~\ref{fig:t14}, 
 for the Planning dimension in Fig.~\ref{fig:t18} and Fig.~\ref{fig:t19}.

\end{document}